\documentclass[journal]{IEEEtran}

\usepackage[OT1]{fontenc}
\usepackage{hyperref}
\usepackage{amsmath,amssymb}     
\usepackage{float}
\usepackage{breqn}
\usepackage[english]{babel}
\usepackage[utf8]{inputenc}

\usepackage[export]{adjustbox}
\usepackage{graphicx}
\usepackage{tabularx}
\usepackage[table]{xcolor}
\usepackage{algpseudocode}
\usepackage{booktabs}

\usepackage[labelfont=bf]{caption}
\usepackage{color}
\usepackage[figurename=Fig.]{caption}
\usepackage{enumitem}
\usepackage{longtable}
\usepackage{makecell}
\usepackage{multirow}
\usepackage{adjustbox}
\usepackage{graphbox}

\usepackage{subcaption}

\usepackage[]{adjustbox}

\begin{document}
	
	\title{ARTxAI: Explainable Artificial Intelligence Curates Deep Representation Learning for Artistic Images using Fuzzy Techniques}
	
	\author{Javier Fumanal-Idocin,~\IEEEmembership{Member,~IEEE}, J. Andreu-Perez,~~\IEEEmembership{Senior Member,~IEEE}, Oscar Cord\'on  ~\IEEEmembership{Fellow,~IEEE},  H. Hagras,~\IEEEmembership{Fellow,~IEEE}, Humberto Bustince,~\IEEEmembership{Fellow,~IEEE}
		
		
		

		
		
		
		
		
		
		
		\thanks{Javier Fumanal-Idocin and Humberto Bustince are with the Departamento de Estadistica, Informatica y Matematicas, Universidad Publica de Navarra, Campus de Arrosadia, 31006, Pamplona, Spain.
			emails: javier.fumanal@unavarra.es, bustince@unavarra.es}
		
		\thanks{Oscar Cordón is with the Dept. of Computer Science and Artificial Intelligence and with Andalusian Research Institute ``Data Science and Computational Intelligence'' (DaSCI), University of Granada, 18071 Granada, Spain. email: ocordon@decsai.ugr.es}
		
		\thanks{Javier Andreu-Perez and Hani Hagras are with the School of Computer Science and Electronic Engineering, University of Essex, Colchester, United Kingdom}
		
		\thanks{Javier Andreu-Perez is with Simbad2, Department of Computer Science, University of Jaen, Jaen, Spain}}

	\markboth{}%
	{Shell \MakeLowercase{\textit{et al.}}: Bare Demo of IEEEtran.cls for IEEE Journals}
	\maketitle
	\begin{abstract}
		Automatic art analysis employs different image processing techniques to classify and categorize works of art. When working with artistic images, we need to take into account further considerations compared to classical image processing. This is because such artistic paintings change drastically depending on the author, the scene depicted, and their artistic style. This can result in features that perform very well in a given task but do not grasp the whole of the visual and symbolic information contained in a painting.
		In this paper, we show how the features obtained from different tasks in artistic image classification are suitable to solve other ones of similar nature. We present different methods to improve the generalization capabilities and performance of artistic classification systems. Furthermore, we propose an explainable artificial intelligence method to map known visual traits of an image with the features used by the deep learning model considering fuzzy rules. These rules show the patterns and variables that are relevant to solve each task and how effective is each of the patterns found. Our results show that our proposed context-aware features can achieve up to $6\%$ and $26\%$ more accurate results than other context- and non-context-aware solutions, respectively, depending on the specific task. We also show that some of the features used by these models can be more clearly correlated to visual traits in the original image than others.
	\end{abstract}

	
	

	\section{Introduction}\label{sec1}
	
	The digitization of numerous paintings and collections worldwide has made it possible to employ the popular computer vision techniques and image processing on artistic data \cite{barni2005image}. One of the most promising topics in this direction is the automatic analysis of paintings, in which these techniques are applied in creative tasks historically performed in most galleries and museums. Some of these are author verification \cite{crowley2016art}, style analysis \cite{lecoutre2017recognizing}, and restoration \cite{zeng2020controllable}. 
	
	Artistic image processing was traditionally performed using hand-crafted or ad-hoc features \cite{carneiro2012artistic}. However, the advent of deep learning and convolutional neural networks has made automatically extracted features very popular \cite{guo2020analysis, tan2016ceci, elgammal2018shape}. Usually, these models are pre-trained and then fine-tuned for each specific task \cite{anwar2018medical, aloysius2017review, rawat2017deep}. This is especially important for the case of artistic images \cite{cetinic2018fine}. One of the actual limitations of these models is that human experts perform their analysis based not only on visual cues but also on their expertise and their knowledge of the historical context, other paintings, materials, etc. \cite{lombardi2005classification}. Adding contextual and historical information to visual cues has been studied to perform different classification tasks in artistic image analysis \cite{garcia2018read, garcia2019context, vaigh2021gcnboost}. However, there is no standard procedure to extract the contextual information associated with each artistic work. Besides, sometimes the context is not encoded in well defined labels. When the information is not well structured, like in a textual commentary, it is also necessary to discriminate those parts relevant to the task. 
	
	One of the most popular approaches to encoding this kind of information is knowledge graphs \cite{garcia2018read, garcia2019context, vaigh2021gcnboost}. A knowledge graph captures the relationships between different concepts and attributions using the structure of a network \cite{taber1991knowledge, chen2020review}. Indeed, graphs are a popular form of representing information \cite{newman2018networks}, and they have been used to solve a myriad of problems in different areas of knowledge, like computer science \cite{Blondel_2008, idocin2020borgia}, biology \cite{palla2005uncovering}, and the social sciences \cite{borgatti2009network, fumanal2021concept}.  However, when using a knowledge graph, a continuous space representation must be constructed from the nodes in the graph. This process is usually performed using deep learning models like node2vec \cite{grover2016node2vec}. Another possibility consists of using multi-task learning, in which a set of different related tasks are trained together so that the information obtained from one is also used in the others \cite{caruana1997multitask}.
	
	Capturing a painting context is also useful for improving the features obtained with a convolutional neural network (CNN). ResNet50 \cite{targ2016resnet} has proven to have good generalization capabilities when trained on the extensively used Imagenet dataset. This generalization capability is particularly important in tasks where there is a significant domain shift and when visual information must be interpreted correctly in order to detect abstract concepts in the image \cite{gonthier2018weakly}. 
	The focus of the current paper is to study how general the features used in an artistic image classification problem are and how useful they are when applied in other similar tasks (i.e. how useful they are to develop transfer learning). Doing so, we also measure if the network is learning a trivial solution to solve the task instead of finding patterns that can be generalized to new observations outside the training set \cite{tonkes2022well}. We also want to test if the features obtained from a black box model can be correlated to known characteristics in the original image.

	In order to achieve these aims, we present different ways to obtain such features, using only visual cues of the image and when additional information is also available. We also propose a new way to represent the contextual embeddings from different paintings using fuzzy memberships that expands previous approaches in this sense \cite{garcia2020contextnet, fumanalArtIPMU}. We shall study how the Fuzzy C-Means clustering algorithm \cite{bezdek1984fcm} and an adapted version of a fuzzy-rule based fuzzy clustering algorithm can be used to construct an embedding space and how this embedding captures relevant information from the original texts. We shall also study the use of Contrastive Language-Image Pre-Training (CLIP) features \cite{conde2021clip, radford2021learning}. In order to map the deep features obtained with these models to known visual cues, we will employ approximate reasoning through the means of fuzzy rules.
	
	This article's \textbf{key innovations} are:
	\begin{itemize}
		\item A \emph{novel} methodology that combines fuzzy clustering and multi-task learning enhances deep learning model performance in artistic image classification.
		\item Extracted deep features are interpreted via fuzzy rules based on semantic information from the original image for {\it  the first time}. We also measure how good is each casual propositional relation as an explanation for each feature.
		\item \emph{Explainable} comparisons between the painting styles of different authors by means of fuzzy rules are developed.
	\end{itemize}
	
	To our mind, this research constitutes an innovative application
	of fuzzy set theory since, to the best of our knowledge, i) fuzzy rules are applied to interpret deep features based on semantic information from the original image; ii) explainable comparisons between authors' painting styles are provided based on fuzzy technologies; and iii) our approach outperforms
	other standard and nonfuzzy computational intelligence techniques in both tasks.

	The rest of the paper is organized as follows: in Section \ref{sec:background} we recall some of the previous concepts required to understand this work and review some relevant literature. In Section \ref{sec:art}, we introduce the proposed framework for artistic image classification using contextual embeddings and the different methods proposed to obtain contextual embeddings from textual annotations. Subsequently, Section \ref{sec:deep_feat} shows the results obtained using the different context-aware and non-context-aware methods. Then, in Section \ref{sec:experimentation}, we discuss the experimental setup and describe our method to explain deep features. Finally, in Section \ref{sec:conclusions}, we give some final conclusions and future lines for this work.
	
	\section{Background} \label{sec:background}
	In this section, we will review some previous works regarding fuzzy clustering and fuzzy rule-based classification, representation learning, and artistic artwork classification.

	
	
	%
	%
	
	
	\subsection{Fuzzy rule-based classification and fuzzy clustering} \label{sec:frbc_algortithm}
	The fuzzy rule-based classification consists of discriminating observations into different categories using rules that follow this structure \cite{cordon1999proposal}:
	\begin{equation}
		\text{IF } \mathbf{x}_1 \text{ is } \mathbf{a}_{j1} \dots \mathbf{x}_n \text{ is } \mathbf{a}_{jn} \text{ THEN class } j \text{ for } j=1,\dots,C
	\end{equation}
	
	\noindent where $\mathbf{x}$ is a multidimensional vector, $j$ is the consequent class, $\mathbf{a}_j$ is an antecedent linguistic value for class $j$, and $C$ is the number of different classes. For this purpose, each attribute is re-escalated into the $[0,1]$ unit interval, and then, the $n$ different attributes are partitioned into different fuzzy subpartitions.
	
	There are different ways in which these fuzzy subpartitions can be generated \cite{CORDON2000187}. There are also different algorithms to generate a set of fuzzy rules to classify the samples \cite{sanz2013ivturs, cord2001genetic}. It is also possible to use fuzzy rules to perform clustering \cite{mansoori2011frbc}. Fuzzy rules have been very useful for explainable AI because they can be easily interpreted by human stakeholders \cite{mendel2021critical}.
	
	Fuzzy C-Means (FCM) is a well known fuzzy clustering algorithm in which each element is assigned not only to one group but rather presents a fuzzy membership to each of the groups considered \cite{bezdek1984fcm}. The algorithm randomly assigns a coefficient for each observation to each cluster. Then, it computes the centroid for each cluster and computes each membership again. The process is repeated until convergence. There is a need to provide a cluster number $c$ as input to the method.

	\subsection{Artistic artwork classification}
	
	Historically, automatic art analysis has been performed using handcrafted features that relied on color, brushwork, and scale-invariant features \cite{carneiro2012artistic, khan2014painting}. Then, once the features were extracted, they were used to train different kinds of classifiers. The most popular tasks include identifying the author, the style, and the theme of a painting \cite{garcia2020contextnet}.
	
	The advent of deep learning has substituted the use of manual features with automatically extracted ones \cite{castellano2022leveraging}. These features have been extracted using different networks, like the Residual Network (ResNet) \cite{targ2016resnet} and the VGG16 \cite{vgg16}. Pre-trained networks can be used to recognize different shapes and entities in images, and they have also been extensively used in art analysis. It is also possible to fine-tune these networks to those tasks \cite{garcia2020contextnet}.
	
	To further study art from a semantic perspective, visual information can be combined with contextual information from the art pieces. There are different ways in which this information can be incorporated into the classification framework. One possibility is to train simultaneously different classification tasks in a multi-task setting \cite{garcia2020contextnet}. In this way, features learned to classify one image can help learning in other tasks and \emph{vice versa}. It is also possible to use knowledge graphs constructed from the dataset itself \cite{garcia2019context} or from external information \cite{castellano2021integrating}.
	
	In spite of these successes, some of these methods still present some shortcomings. It is difficult to tell when the predictions are based on meaningful artistic knowledge of shortcuts. Besides, the categorization of specific parts of an image (like a cat, a dog, etc.) differs from aspects like an author, which cannot be necessarily correlated only to specific parts of the image. This makes some explainability models like Grad-CAM \cite{selvaraju2017grad} less useful for this task.

	\section{Proposed Artistic Image Classification Framework} \label{sec:art}
	
	\begin{figure*}[ht]
		\centering
		\includegraphics[width=\linewidth]{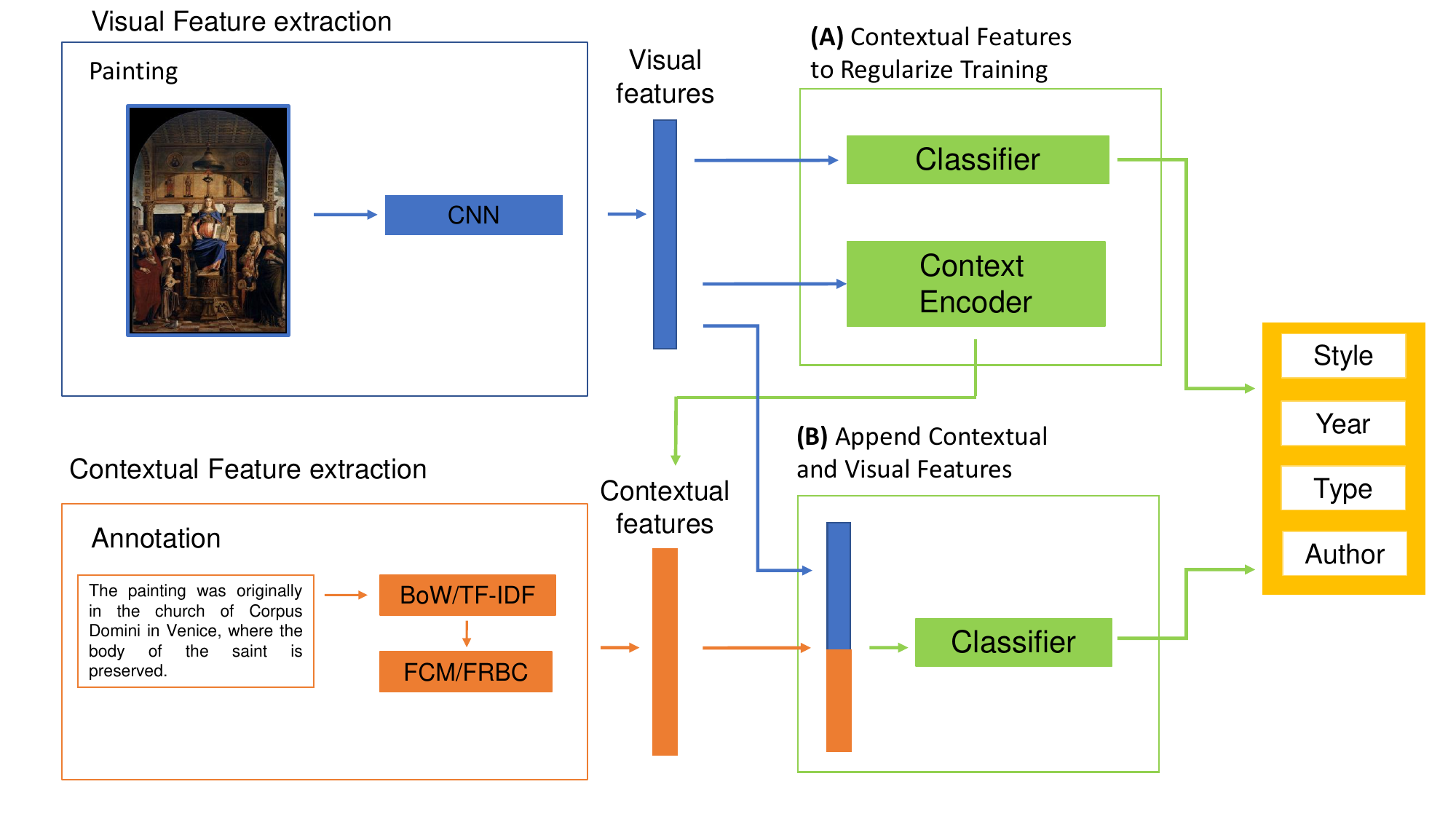}
		\caption{\textbf{Scheme of the two proposed classification frameworks using contextual features.} Option \textbf{(A)} uses the contextual features to regularise the training of the CNN. The last layer of the network is replaced to fit the number of classes in the task. Option \textbf{(B)} appends the visual and contextual features in a single vector that is fed to a final classification layer.}
		\label{fig:scheme}
	\end{figure*}

	In this section, we shall discuss our proposed classification framework, illustrated in Fig. \ref{fig:scheme}, and the different alternatives studied to train and finetune such a model. This model shall be capable of solving one task or various tasks at the same time (style, year, type, and author identification) at the same time.
	
	\subsection{Model architecture} \label{sec:model_arch}
	
	Our proposed framework consists of two different parts (see Fig. \ref{fig:scheme}). We used a ResNet50 \cite{targ2016resnet} to extract visual features for each image. The last layer of the ResNet is substituted by a linear classifier with the appropriate number of classes as the output.
	
	In order to train the model, we have used different loss strategies. The most basic strategy is to train the model using the features obtained from ResNet using the standard cross-entropy loss for all classes $C$: 
	
	\begin{equation}
		l(y, \hat{y}) = - \frac{1}{N}\sum_{i=1}^{N}\sum_{c=1}^{C} y_{ci} \log \hat{y}_{ci} + (1 - y_{ci}) \log (1 - \hat{y}_{ci})  
	\end{equation}

	In order to exploit all the information available in the training process for each task, we train a combination of losses in a multi-task learning (MTL) setting, fusing them in a single simultaneous training objective for the author, type, school, and timeframe. In that case, the classification loss is the average of the four different cross-entropies for each task:
	
	\begin{dmath}l_{MTL} = \frac{1}{4} \left( l(y_{author}, \hat{y}_{author}) + l(y_{type}, \hat{y}_{type}) + l(y_{school}, \hat{y}_{school}) + l(y_{time}, \hat{y}_{time}) \right)
	\end{dmath}
	
	\subsection{Context extraction}

	In order to aid regularisation in the classification task, we use additional contextual information to condition the output from our model. This information is encoded using a real-valued vector, computed using one of the methods proposed:
	
	\begin{enumerate}
		\item Node2vec in a knowledge graph that connects the paintings according to their shared attributes \cite{garcia2020contextnet}.
		\item Fuzzy memberships over the contextual annotations for each painting.
		\item Features obtained with a CLIP autoencoder on the contextual annotations.
	\end{enumerate}
	
	Concerning the second method, {\it fuzzy context encoding}, the idea of using fuzzy clustering for this task is that the space formed using a word embedding method can be a faithful representation of the original domain, but it might not be useful to solve the task at hand. Since we are interested in using these features to discriminate between classes, we are more interested in the topology of the representation obtained and the groups that are naturally present in them.
	
	We expect that these groups agglomerate categories that are not mutually exclusive. For example, in the case of artistic representation, style and year can be very correlated because of artistic movements. In this case, there are many more possible examples: landscapes can be grouped together but belong to different authors, etc. Fuzzy clustering is the most suitable clustering tool for this task since we intend to express the membership to different, not mutually exclusive groups. For each observation, we have a fuzzy membership degree for each pertinent group. 
	
	Besides, fuzzy clustering is much more convenient than the traditional K-means algorithm for this task since fuzzy memberships are all in the same $[0,1]$ range. This means that the resulting vector will be a real-valued vector with more information than a one-hot encoding corresponding to a unique cluster. In addition, memberships to clusters that are far away can be modeled with a $0$ in both cases, while distances in the K-means can be very different in magnitude but represent the same thing: the observation is not in this cluster. Thus, fuzzy memberships are also more suitable than Euclidean distances in a hard clustering approach for this task.
	
	In order to compute the fuzzy memberships, we first encode the text annotations using Bag of Words (BoW) or  TF-IDF encodings. 
	Once this codification is computed, we run a fuzzy clustering algorithm to obtain the desired number of memberships. 
	
	The most popular fuzzy clustering algorithm for raw observations is the FCM \cite{bezdek1984fcm} to obtain soft partitions of the feature domain. Fuzzy rules have also been intensively used to perform classification and data mining. However, many clustering algorithms are based on the direct optimization of a partitioning objective function without intermediate structures that explain the rationale of the decision for the optimal partition solution. Hence, we adopt the approach proposed by Mansoori et al. of integrating a fuzzy rule-based process inference in the clustering formation \cite{mansoori2011frbc}.
	The modification of the original clustering algorithm includes two main changes:
	
	\begin{enumerate}
		\item The original algorithm gives, as a result, a crisp clustering. In order to return fuzzy memberships to the distinct groups, we use the value of the consequent for each rule selected in the algorithm.
		\item The original algorithm had a stopping condition based on a stopping parameter so that when a percentage of the original data was assigned to a group, it ends its execution. Since there are no proper criteria to choose this parameter, we stop when all the original samples have been removed from the dataset.
	\end{enumerate}
	The resulting algorithm is the following:
	\begin{algorithmic}
		\State Set $X$ as the main data observations: $X \gets {\mathbf{x}_1, \dots, \mathbf{x}_m}$
		\State Set $z$ to $1$
		\Repeat
		\State Generate a set of synthetic samples $X'$
		\State Compute the distance of the main data $X$ to its centroid $q$, and compute the distance of the synthetic data $X'$ to its centroid $q'$
		\If{$q' < q$}
		\State Go to the start of the loop
		\EndIf
		\State Generate a set of rules to discriminate between $X$ and $X'$
		\State Select the rule with the highest average value for the antecedent
		\State The membership of each observation in cluster $z$ is the degree of truth for the consequent of the chosen rule
		\State Remove the observations from $X$ and $X'$ that belonged to cluster $z$ with a degree greater than $0.5$
		\State Increment $z$ by $1$
		\Until{$|X| = 0$}
	\end{algorithmic}	
	
	This approach has an advantage over the FCM. In the FCM, the sum of all memberships must sum $1$. This means that the bigger the contextual vector is, the lesser value each membership will be. In the case of fuzzy rule-based clustering, memberships for each cluster are independent, so there is no such restriction. In this way, it works as a possibilistic FCM algorithm \cite{1492404}. Furthermore, we do not need to specify the number of clusters, as the process has a natural way to finish when all the observations have been assigned to a group.
	
	
	\subsection{Context conditioning}
	
	Once we have computed the contextual vector, there are two different approaches to combine it with the visual cues:
	
	\begin{itemize}
		\item We use the contextual information vector in order to ``regularise'' the visual features. To do so, we have two ``final'' layers: one encoder that transforms the final feature vector of the network into the contextual features and another one that performs the classification.  These encoders are single fully connected layers with a Rectified Linear Unit activation function (Fig. \ref{fig:scheme}a). This is a trendy scheme in the literature to join information from heterogeneous sources \cite{radford2021learning}.
		\item We append the contextual information vector to the visual characteristics vector. Then, we use a fully connected layer to learn from the resulting vector (Fig. \ref{fig:scheme}b).
	\end{itemize}

	Given $r$, the final embedding obtained from the ResNet, and $m$ the number of clusters obtained with the fuzzy clustering, the loss function for the reconstruction of the fuzzy membership vector is the Smooth L1:
	
	\begin{equation}
		\delta_{emb}(a, b) = \begin{cases}
			\frac{1}{2}(a - b), & \text{if}\ |a - b| \leq 1 \\
			|a - b| - \frac{1}{2}, & \text{otherwise}
		\end{cases}
	\end{equation}
	
	\begin{equation}
		l_{emb} = \sum_{i=1}^{n} \sum_{j=1}^{m} \delta_{emb}(w_{ij}, r_{ij})
	\end{equation},
	
	\noindent where $a,b \in \mathbb{R}$, $w_{ij}$ is the $j$-th element of the fuzzy membership of the $i$-th sample,  $r_{ij}$ is the $j$-th element of the reconstructed contextual vector of the $i$-th sample.
	
	Finally, we combine both the classification and the embedding loss using a convex combination of both, so the final loss is:
	
	\begin{equation}
		l = \alpha l_{class} + (1 - \alpha) l_{emb},
	\end{equation}
	
	\noindent	where $\alpha \in [0,1]$. Choosing the correct $\alpha$ value is important in order to avoid one loss function ``dominate'' the other. We have chosen $0.9$ as the value since it gave good results in the literature \cite{garcia2019context}.



	\section{Explaining Deep Features using Fuzzy rules} \label{sec:deep_feat}
	
	The interpretation of deep features in CNNs is still a challenging task, as they are the output of a large number of matrix operations. It is possible to grasp a better understanding of their behaviour if we correlate them to known characteristics in the original image. 
	
	In order to do so, we propose to use a FRBC that will map known features to the degree of activation of the deep features used to classify the paintings in each task. This will allow us to understand the predictions done by the network using abstract concepts  and Grad-CAM heatmaps \cite{selvaraju2017grad}. 
	
	\subsection{Extracting style information}
	To extract known features, we first use another ResNet50 model trained to recognize artistic movements in a painting. The output of this model will be a vector containing the score  for all possible styles considered. This model is trained on $100,000$ paintings belonging to $2,300$ different authors from the WikiArt dataset, which contains $27$ different artistic styles. We trained the ResNet for a maximum of 300 epochs with a limit of $50$ iterations without a tangible improvement in the model performance. It obtained a $53\%$ accuracy overall (specific details about the datasets used can be found in Section \ref{sec:datasets}).

	\subsection{Characterizing visual focus}
	\label{sec:gradcam}
	
	In order to join conceptual and visual concepts, we have studied the gradient maps of the SemArt models using Grad-CAM \cite{selvaraju2017grad}, which shows the regions which contributed significantly to the network prediction. Since we are studying four different tasks, we obtain for each image not one, but four different Grad-CAM heatmaps. To get overall information we fuse them using the average of those values. Once we reduced the different Grad-CAM maps to only one, we characterize each of them focusing on three different properties:
	
	\begin{enumerate}
		\item The percentage of the image with significant attention values.
		\item The magnitude of the biggest gradient in the heatmap.
		\item Explanation of connected parts in the heatmap.
	\end{enumerate}
	
	In order to consider a pixel relevant, we just compare it against the average value for that image. Those bigger than the average are considered relevant.
	
	In order to compute the maximum gradient in the image, we compute the Sobel filter, both in the horizontal and vertical axis. Once we have the gradient for each direction, we compute the gradient magnitude in each point from those vectors using the Pythagorean theorem. Then, we choose the biggest one as a result.
	
\begin{figure}
		\centering
		\includegraphics[width=\linewidth]{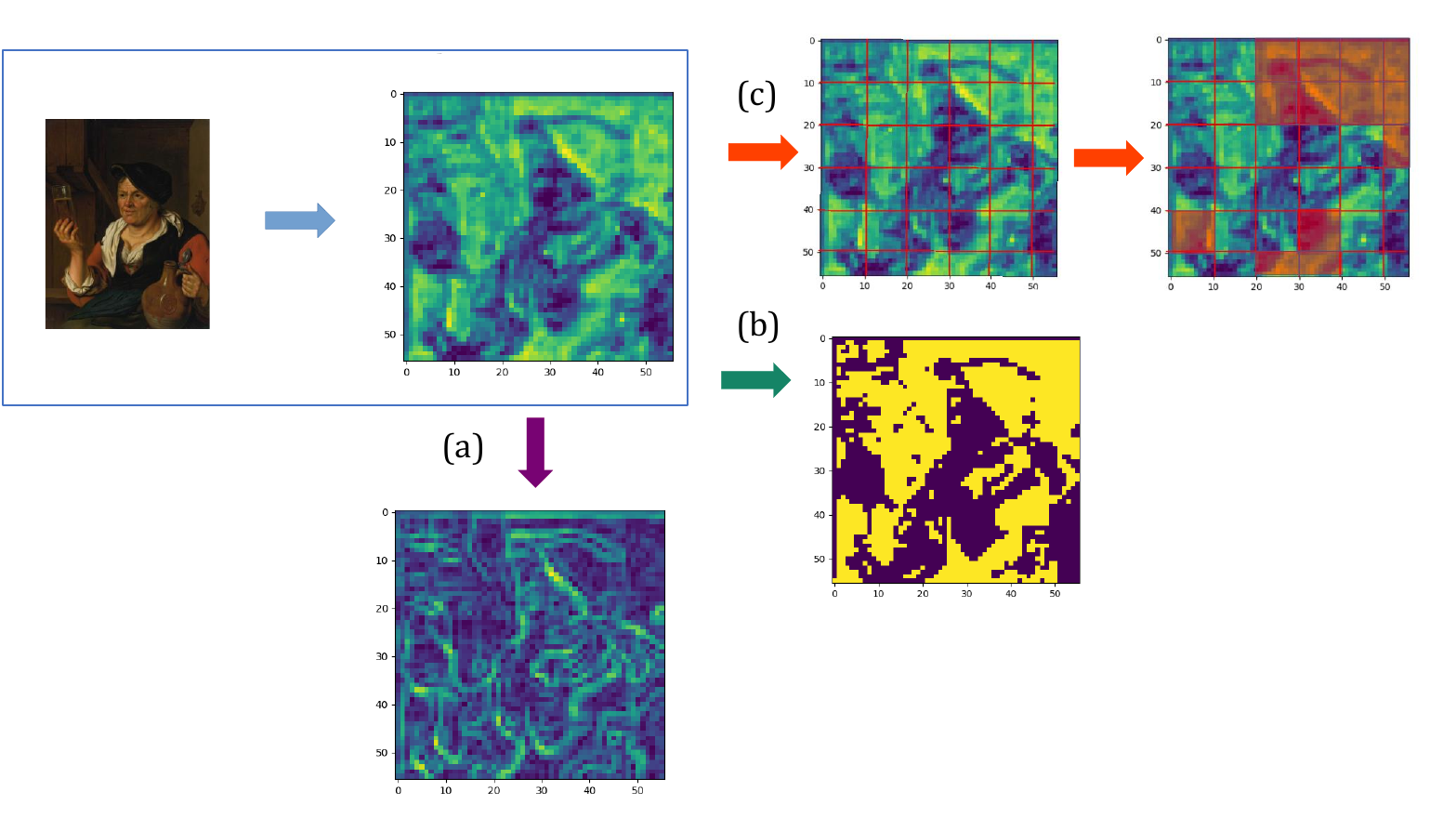}
		\begin{tabular}{cc}
			\toprule
			Descriptor & Value\\
			\midrule
			Maximum gradient  &  1.8 \\
			Relevant area  &  0.53\\
			Number of super regions  & 3 \\
			\bottomrule
		\end{tabular}

		\caption{\textbf{Example of the extraction of the different descriptors for a Grad-CAM heatmap.} (a) shows the gradient magnitude for each pixel. We choose the highest value from that image as the descriptor. (b) shows the pixels denoted as relevant because their value was higher than the average value in the image. (c) shows the division into different regions and those designed as relevant. Then, we can generate the ``super'' regions. In the adjacent table, we can find the numeric values for each descriptor in this image.}
		\label{fig:example_gradcam_descriptor}
	\end{figure}

	We divide the image into $N$ squares of equal size to designate the number of connected components in the image. Then, we denote which of these regions was relevant in the classification. We designate as relevant those regions whose average value is bigger than the average value of the whole image. Then, we connect the regions regarded as relevant that are adjacent, which results in a series of ``super'' regions. The number of connected components obtained is the same as the number of the ``super'' regions formed in the image. Fig. \ref{fig:example_gradcam_descriptor} shows an example of the results for this characterization for one image.
	
	Finally, Table \ref{tab:descriptors} shows a summary of the statistics of these descriptors. We can see that the average value of the relevant parts of the image is about a third and that there is an average of 3 connected components in each painting.
	
	
	\begin{table}[]
		\centering
		\caption{Statistics of the Grad-CAM heatmap descriptors for the SemArt datase.}
		\begin{tabular}{ccc}
			\toprule
			Descriptor & Average value & Standard deviation\\
			\midrule
			Maximum gradient  &  2.27  & 0.49 \\
			Relevant area  &  0.37 & 0.08\\
			Number of super regions  & 2.90 & 1.39 \\
			\bottomrule
		\end{tabular}
		
		\label{tab:descriptors}
	\end{table}

	\subsection{Mapping known features to deep features} \label{sec:my_frbc}
	
	In order to map interpretable patterns from the known features to others, we use a FRBC. Our aim in using the FRBC is to obtain an interpretable classifier. In order to do so, we set $15$ as the maximum number of rules and $4$ as the maximum number of antecedents. However, the real number of rules used will be reduced from that number using a quality metric. We also choose three linguistic labels for the fuzzy partitions that are easily interpretable: low, medium, and high. To obtain the prediction for a sample, we compute the dominance score (DS) of each rule $r$ \cite{kiani2022temporal}. This metric measures how often the rule fires and how its strength is compared to the remainder so that rules that are good in both senses are preferred. 
	
	The DS for each rule is the product of their support and confidence, with the support being \cite{andreu2021explainable}:
	
	\begin{equation} \label{eq:support}
		s_r(Ants_r \rightarrow Cons_r) = \frac{\sum_{\mathbf{x} \in (Ants_r \rightarrow
				Cons_r)} w_r (\mathbf{x})}{|R|},
	\end{equation}
	
	\noindent where $w_p(x)$ is the degree of truth of the antecedents of rule $r$ for the sample $x$ and $R$ is the set of all rules in the FRBC. The firing strength of the rule is the product of the truth degrees of all antecedents of the rule (product t-norm). Confidence is defined as:
	\begin{equation}
		c_r(Ants_r \rightarrow Cons_r) = \frac{\sum_{\mathbf{x} \in (Ants_r \rightarrow
				Cons_r)} w_r (\mathbf{x})}{{\sum_{r=1, \mathbf{x} \in (Ants_r) }{w_r(\mathbf{x})}}}
	\end{equation}.
	
	So, the DS of each rule, $ds_r$, is defined as:
	
	\begin{equation}
		ds_r =  s_r * c_r
	\end{equation}
	
	Finally, we compute the association degree, $as_r(x)$, using $w_r(\mathbf{x})$ and $ds_r$:
	\begin{equation}
		as_r(x) = w_r(\mathbf{x}) * ds_r.
	\end{equation}
	
	Each sample is classified according to the consequent class of the rule with the highest association degree for that sample:
	
	\begin{equation} \label{eq:rule_pred}
		P(x) = Cons_{arg\,max({as_r(x) \forall r \in R})}
	\end{equation}

	For our experimentation, we have trained a FRBC. In order to train one, we used a genetic algorithm that optimizes the fuzzy partitions and the antecedents and consequents for each rule. The metric to optimize is the Matthew correlation coefficient (MCC):
	\begin{equation}
		MCC = \frac{(TP \times TN) - (FP \times FN)}{\sqrt{(TP + FP)(TP + FN)(TN + FP)(TN + FN)}}
	\end{equation}
	where TP is true positive, TN means true negative, FP is false positive, and FN is false negative.
	
	We also add another condition: in case two subjects performed equally on the fitness metric, we prefer those that did so using fewer rules.
	
	
	\section{Experimentation and results} \label{sec:experimentation}
	In this section, we evaluate the performance of the proposed framework in the artistic image classification problem for three different settings:
	
	\begin{enumerate}
		\item Author identification using fuzzy rules.
		\item Classification of type, school, timeframe, and author for paintings using single and multi-task settings.
		\item Remaining tasks deep features activations using fuzzy rules.
	\end{enumerate}
	
	\subsection{Datasets} \label{sec:datasets}
	
	For our experimentation, we have used the SemArt dataset \cite{garcia2018read}. This dataset consists of 21,384
	painting images. Following the original data partition in \cite{garcia2018read}, 19,244 images are used for training (i.e., a $90\%$), $1,069$ for validation, and $1,069$ for test (i.e., a $5\%$ each). Each painting has an associated textual artistic comment. In this dataset four different classification tasks are proposed:
	
	\begin{itemize}
		\item Type: each painting is classified according to $10$ different common types of paintings: portrait, landscape, religious, etc.
		\item School: each painting is identified with different schools of art: Italian, Dutch, French, Spanish, etc. There are a total of $25$ classes of this kind.
		\item Timeframe: this attribute, which
		corresponds to periods of 50 years evenly distributed between 801 and 1900, is used to classify each painting according to its creation date. We consider only the timeframes where more than $10$ paintings are present. This corresponds to $18$ classes.
		\item Author: it corresponds to the author of each painting. We consider a total of $350$ painters, that comprise the set of authors with more than $10$ paintings in the dataset.
	\end{itemize}
	
	
	We have also used the WikiArt dataset. This dataset is a collection of high-resolution images of artworks and their associated metadata that were scraped from Wikipedia \cite{saleh2015large}. The WikiArt dataset contains over 81,000 images of fine art paintings representing a wide range of artistic styles and historical periods from the 11th century to the present day. Each image in the dataset is accompanied by a set of metadata, including the title of the artwork, the artist, the year of creation, the medium used, and the dimensions of the artwork, among other attributes.

	\subsection{Results for mapping known features to discriminate particular authors}

	We will first consider author identification, which is one of the most relevant tasks of artistic curation. Not only to properly identify the original painter of one artistic piece but also to detect possible forgeries or false attributions. Deep learning models can output the likelihood of a painting's authorship, but these predictions tend to be overconfident and unrealistic. In order to solve this problem, we use a FRBC to distinguish between particular authors of interest and output a more trustworthy estimation of the reliability of such prediction \cite{9786614}.
	
	As a way of illustrating this application, we shall construct a FRBC to distinguish two painters that were acquainted in real life: Paul Gauguin and Vincent Van Gogh. It is exciting to see if the resulting rules match the actual knowledge that we have of both of them. They are considered post-impressionist painters, with strong similarities and differences in their style. 
	
	First, to obtain the style characterization of the SemArt paintings, we apply the style characterization model trained on the WikiArt dataset. We do not expect a significant domain shift between both datasets as the SemArt dataset was not collected with a particular bias in the selection process. However, in order to check how good the performance was in SemArt, we compared the results using the painters that have paintings in both datasets. We measured how often Semart assigned a style to a painting that was different from the styles that Wikiart listed for the painter (Table \ref{tab:authors_style_validation}). We found a total of 8 common artists in both datasets, with different degrees of misclassification. Of course, some errors are more important than others, i.e., it is not the same to incorrectly classify a pointillist painting as impressionist rather than medieval art. However, the whole complexity of this problem is left open in this work, which is closely related to ordinal classification problems \cite{9186013}. We found the results satisfactory, as only one painter found significant missclasification: Albert Durer. The rest presented an error of less than 0.10 or included very few paintings (Singer Sargent).

	\begin{table}
		\centering
		\caption{Results for author and style correlation in Semart. Predictions for each style are generated on a ResNet fine-tuned in the Wikiart dataset. The incorrect style indicates how many times a painting was assigned to a style that does not correspond to the author by the model.}
		
		\begin{tabular}{ccc}
			\toprule
			Artist & Number of paintings in Semart & Incorrect style \\
			\midrule
			Albrecht Durer      & 79 &   0.25 \\
			Camille Pissarro    & 22 &   0.00 \\ 
			Childe Hassam       & 8 & 0.00  \\ 
			Claude Monet        & 92 &  0.03  \\
			Edgar Degas         & 64&  0.03 \\
			John Singer Sargent & 6& 0.16   \\
			Paul Cezanne        & 76 &  0.07  \\
			Vincent Van Gogh    & 291& 0.04  \\
			\bottomrule
		\end{tabular}
		
		\label{tab:authors_style_validation}
	\end{table}
	
\begin{figure}
		\centering
		\subfloat[]{\includegraphics[width=.45\linewidth]{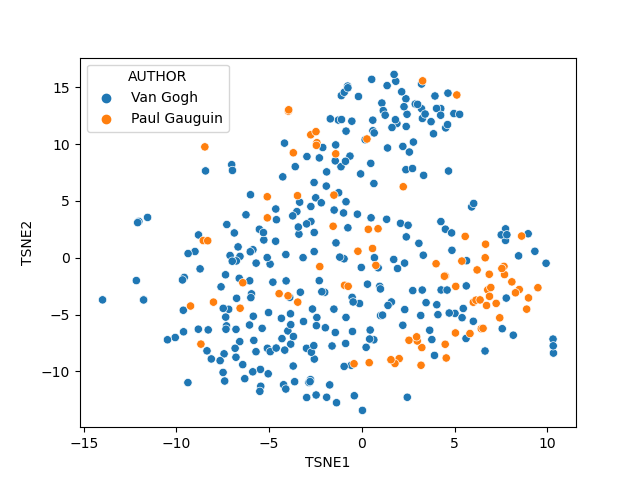}}
		\subfloat[]{\includegraphics[width=.45\linewidth]{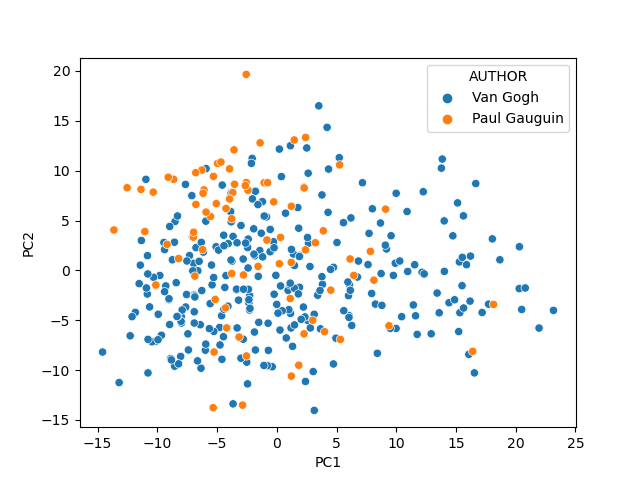}}
		\caption{\textbf{Visualisation of Van Gogh and Gauguin samples} using TSNE (a) and PCA (b). The features reduced are the style and Grad-CAM features.}
		\label{fig:gogh_pca}
	\end{figure}

	
	The SemArt dataset contains 291 paintings from Van Gogh and 81 from Gauguin. Fig. \ref{fig:gogh_pca} shows a PCA and a t-distributed stochastic neighbor embedding (TSNE) visualization of the paintings for each class using our style and Grad-CAM characterization. We derive the rules tuning a genetic algorithm for the FRBC described in Section \ref{sec:my_frbc}. As the number of possible antecedents is too high for the genetic algorithm to obtain a good result, we first trained a Gradient boosting classifier (GBC) \cite{chen2015xgboost} using all the input features. This classifier obtained a $100\%$ accuracy but used many features that we know should not be relevant to this task. Then, we computed the feature importance for this classifier and used the seven most important ones to train the FRBC. Table \ref{tab:performance_van_gogh} shows the performance for both the GBC and the FRBC on the selected set of features. We obtained good results for both approaches, which again proves that the style and Grad-CAM heatmap characterization was successful. Indeed, we obtained better results with the FRBC than the GBC in the reduced set of features.
	
\begin{figure*}[]
		\centering
		
		\adjustbox{max width=\linewidth, center}{
			\begin{tabular}{ccl|ccc}
				\toprule    
				Author & & Antecedents & DS & Train Acc & Test Acc\\
				\midrule
				\multirow{5}{*}{\includegraphics[width=.10\linewidth]{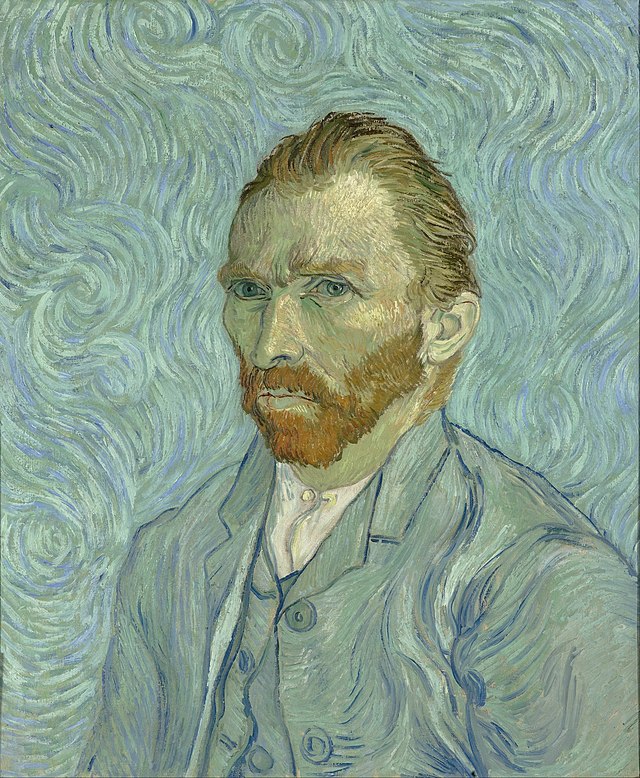}}
				& \cellcolor{gray!25} 1 & \cellcolor{gray!25}IF New Realism IS Low AND Post Impressionism IS Medium & \cellcolor{gray!25}0.0076 & \cellcolor{gray!25}0.5000 & \cellcolor{gray!25}0.0000 \\
				& 2 & \textbf{IF Early Renaissance IS Medium AND New Realism IS Medium AND Synthetic Cubism IS Medium} & 0.0740 & 0.7777 & 1.0000 \\
				& \cellcolor{gray!25} 3 &  \cellcolor{gray!25}\textbf{IF Early Renaissance IS Low AND Synthetic Cubism IS High} & \cellcolor{gray!25} 0.2517 & \cellcolor{gray!25} 0.9390 & \cellcolor{gray!25} 0.8888 \\
				& 4 & \textbf{IF Synthetic Cubism IS Low} & 0.4624 & 0.9097 &  0.9250 \\
				& \cellcolor{gray!25} 5 &  \cellcolor{gray!25}IF Contemporary Realism IS Medium AND Synthetic Cubism IS Low AND Relevant area IS Low & \cellcolor{gray!25} 0.0092 & \cellcolor{gray!25} 0.0000 &  \cellcolor{gray!25} 0.0000\\
				&&&&& \\
				\midrule
				\multirow{2}{*}{\includegraphics[width=.10\linewidth]{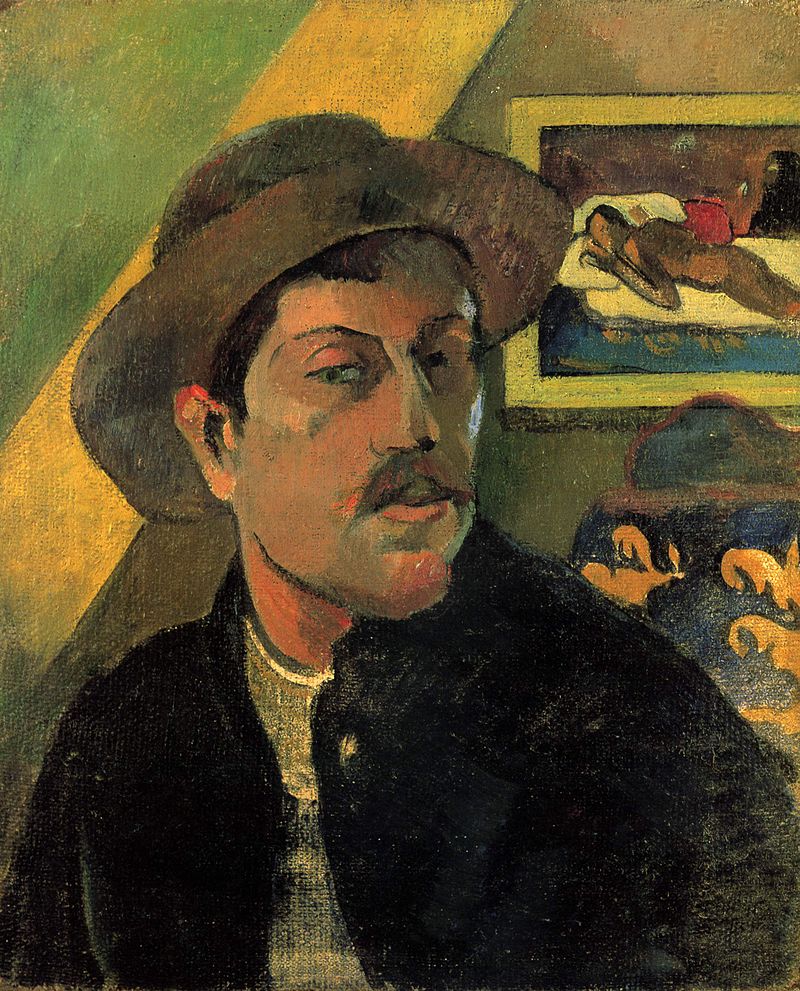}}  
				
				& 6 & \textbf{IF Contemporary Realism IS Medium AND Minimalism IS Low} & 0.3389 & 0.7586 & 0.7692\\
				& \cellcolor{gray!25} 7 &  \cellcolor{gray!25}IF Early Renaissance IS Medium AND Minimalism IS Medium AND Synthetic Cubism IS Medium & \cellcolor{gray!25} 0.0124 & \cellcolor{gray!25} 0.0000 & \cellcolor{gray!25} 0.0000 \\
				&&&&& \\
				&&&&& \\
				&&&&& \\
				&&&&& \\
				\bottomrule
				
		\end{tabular}}
		\caption{Rules that differentiate Van Gogh from Gauguin paintings.}
		\label{tab:gogh_gauguin}
	\end{figure*}

	Fig. \ref{tab:gogh_gauguin} shows the rules obtained to differentiate both authors and the DS and individual accuracy for each rule in all the training samples they fired.  We obtained three successful rules for Van Gogh and one for Gauguin. From those rules, we can see that Synthetic Cubism is a very good feature to identify Van Gogh's paintings compared to Gauguin, and the Early Renaissance style is the second best. We only found one relevant pattern for Gauguin. One interesting issue is that the best features to discriminate both artists are styles that did not exist in the actual time of the painters. This can indicate that these painters already started some of the traits that characterized those artistic movements.

	
	
	\begin{table}[]
		\centering
		\caption{Performance for a GBC and our FRBC in the Van Gogh/Gauguin identification task using the relevant features found from a GBC on the whole set of features.}
		\begin{tabular}{cccc}
			\toprule
			\multicolumn{2}{c}{GBC} & \multicolumn{2}{c}{FRBC} \\
			Accuracy & MCC & Accuracy & MCC \\
			\midrule
			0.88 & 0.53 & 0.88 & 0.62 \\
			\bottomrule
		\end{tabular}
		
		\label{tab:performance_van_gogh}
	\end{table}
	
	\subsection{Results for the classification tasks} \label{sec:methods}
	
	We studied the four different classification tasks (type, school, timeframe, and author identification) using the training/test partitions as described in Section \ref{sec:datasets}. To measure the performance for each task, we have computed the classification accuracy.  using The following classification methods are considered:
	
	\begin{enumerate}
		\item The ResNet50 and the VGG16 networks using their corresponding pre-trained weights. We adapt the last layer to match the number of target classes. These solutions only consider the visual information for each image.
		
		\item The ResNet50 and the VGG16 fine-tuned in an MTL setting for all the different classes, so that context is captured by the shared information between tasks. We retrained all the weights for each network.
		
		\item ContextNet: The ResNet50 with information captured from contextual annotations and metadata, using node2vec representations represented by a knowledge graph (KGM) \cite{garcia2019context}.
		
		\item Our proposed classification framework in Fig. \ref{fig:scheme} uses the ResNet50 to extract visual features and BoW/TF-IDF and FCM to encode the textual annotations. We use the contextual features as a regularisation element in the training process and append both vectors of features (marked as ``append'' in Table \ref{tab:res}). For the case of the BoW codification, we also test a lighter model that uses only the top $100$ most popular words.
		
		\item Our proposed classification framework in Fig. \ref{fig:scheme} uses the ResNet50 to extract visual features and BoW/TF-IDF and FRBC to encode the textual annotations. We use the contextual features as a regularising element in the training process and append both vectors of features (marked as ``append'' in Table \ref{tab:res}).
		
		\item Our proposed classification framework in Fig. \ref{fig:scheme} uses the ResNet50 to extract visual features and a CLIP autoencoder to encode the textual annotations.
		
		\item The combination of our MTL approach and contextual encodings using FCM and CLIP.
		
	\end{enumerate}
	
	Table \ref{tab:res} shows the results for each of the tasks and models. Analyzing from top to bottom, we can see that the MTL methods alone (i.e., only based on visual information) performed worst than those that used a KGM or FCM to capture contextual information. Comparing the KGM and FCM encodings, the BoW-FCM performed better in all tasks but ``Author''. When considering contextual information using MTL, KGM, and FCM-based methods, the performance improved substantially for all classes. 
	
	The best result for each different task was obtained using an FCM-based model in all cases but TimeFrame. However, those that used most words for the contextual embedding performed poorly on the Author task, in which the BoW model with only $100$ words performed significantly better than the rest of the FCM-based models. It was also the best performing method for this class compared to the KGM and MTL-based proposals. This could be due to the fact that the ``Author'' classification is the most complicated task, with only a few learning examples per class, and the available contextual vectors are not specific enough to help discriminate in those cases. Appending the contextual vector instead of using it to regularise the gradient seems to have a similar effect on the final performance of the system. Since we are not guaranteed to have textual annotations, it is preferable to use those that only required them in the training process.
	
	Finally, we have also joined both paradigms using an MTL model with the two different contextual vectors as a regularisation element. These models outperformed the rest of the models considering the average of all tasks. MTL-FCM and MTL-CLIP obtained an average of 0.661 and 0.662 accuracy, respectively, whilst the second best model, the BoW$_{100}$ + FCM, obtained a 0.647. The best existing previous approach, the ContextNet, obtained the fourth best performance with $0.6444$ average accuracy on the four tasks.

	\begin{table}
		\centering
		\caption{Correct Classification Ratio results for the different attributes on SemArt Dataset test partition (see  Section \ref{sec:model_arch} and Section \ref{sec:methods} for more details).}
		\adjustbox{width=\linewidth}{
			\begin{tabular}{cccccc|c}
				\toprule
				& Method & Type & School & TimeFrame & Author & Average\\
				\midrule
				\multirow{2}{*}{1)} & VGG16 & 0.706 & 0.502 &  0.418& 0.482 & 0.527\\
				& ResNet50 & 0.726 & 0.557 & 0.456 & 0.500 & 0.559\\
				\midrule
				\multirow{2}{*}{2)} & VGG16 MTL & 0.732 & 0.585 & 0.497 & 0.513 & 0.581  \\
				& ResNet50 MTL & 0.763 & 0.565 & 0.464 & 0.431 & 0.555 \\
				\midrule
				\multirow{1}{*}{3)} & ContextNet & 0.786 & 0.647 & 0.597 & 0.548 & 0.644	 \\
				\midrule
				\multirow{5}{*}{4)}& BoW + FCM& 0.794 & 0.655 & 0.604 & 0.238 & 0.572 \\
				& BoW + FCM-apppend & 0.802 & 0.654 & 0.584 & 0.230 & 0.567\\
				& TF-IDF  + FCM & 0.786  & 0.645  & 0.604 &  0.229 & 0.566\\
				& TF-IDF  + FCM-append & 0.778 & 0.654  & 0.589 &  0.226 & 0.561 \\
				& BoW$_{100}$ + FCM & 0.792 & 0.630 & 0.586 & \textbf{0.559} & 0.647 \\
				\midrule
				\multirow{2}{*}{5)} & TF-IDF + FRBC & 0.785 & 0.643  & 0.597 & 0.233 & 0.564\\
				& TF-IDF + FRBC-append & 0.759 & 0.623 & 0.533 & 0.154 & 0.517 \\
				\midrule
				\multirow{1}{*}{6)} & CLIP-context &  0.784 & 0.649   & 0.601 & 0.215 & 0.560 \\
				\midrule
				\multirow{2}{*}{7)} & MTL-FCM & \textbf{0.804} & \textbf{0.691}  & 0.618  & 0.531  & 0.661 \\
				& MTL-CLIP & 0.790 & 0.677 & \textbf{0.630} & 0.551 & \textbf{0.662} \\

				\bottomrule
			\end{tabular}
		}
		\label{tab:res}
	\end{table}
	
	\subsection{Results mapping style characteristics to deep features}

\begin{figure}[h]
		\centering
		\subfloat[][]{\includegraphics[width=.45\linewidth]{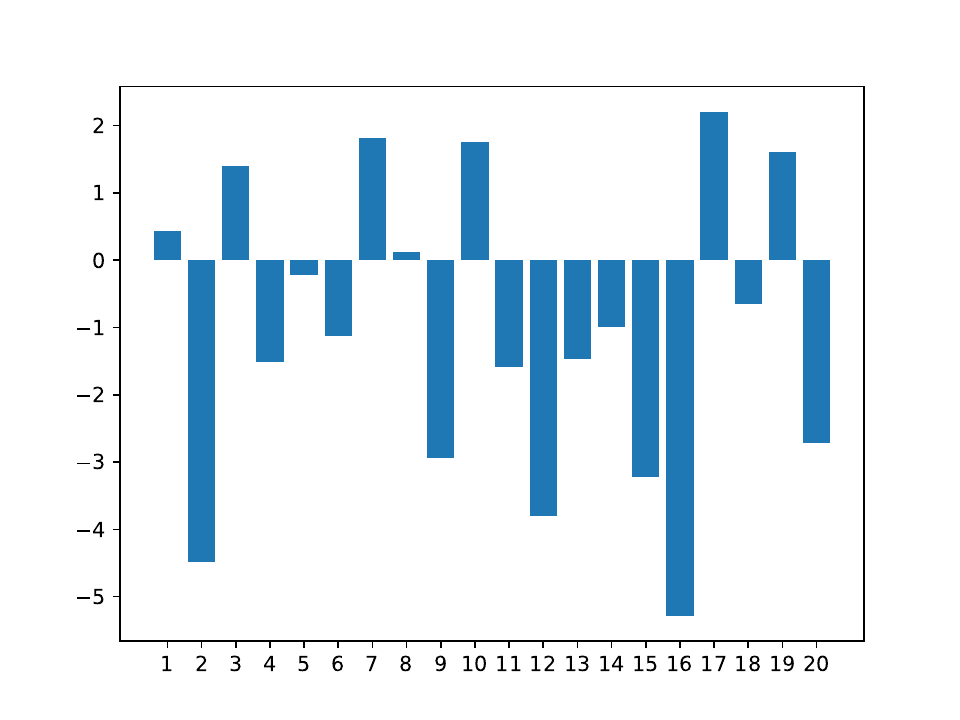}}
		\subfloat[][]{\includegraphics[width=.45\linewidth]{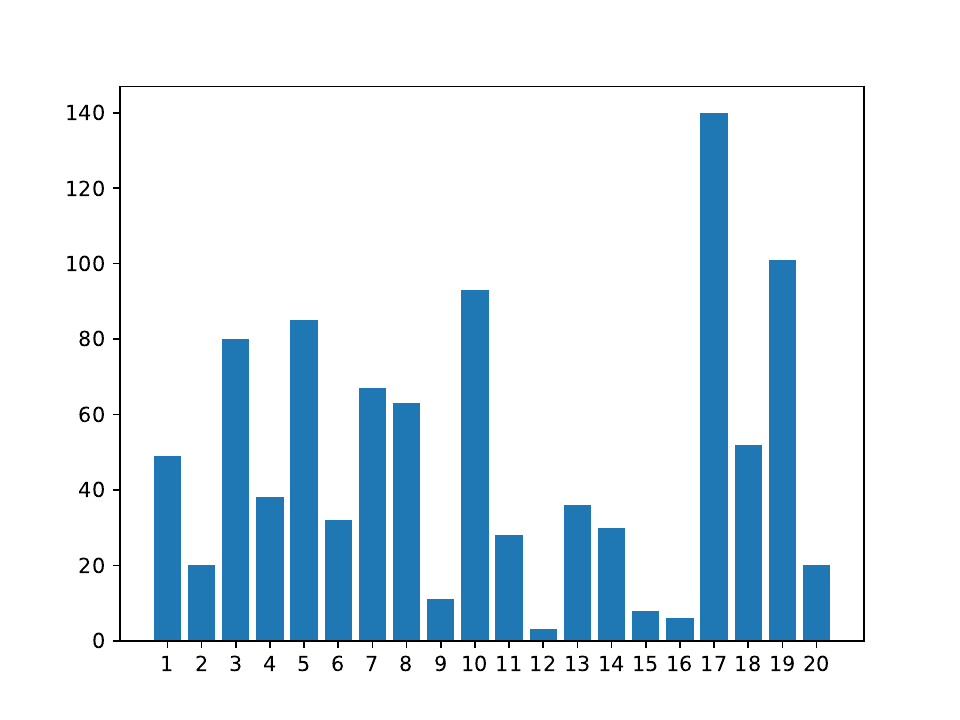}}
		
		\caption{\textbf{Study of deep feature activations.} (a) Average value of the 20 deep features used in MTL-FCM predictions. (b) Histogram containing the number of times that each feature presented the biggest value for each sample in the training set.}
		\label{fig:features_deep_plot}
	\end{figure}

\begin{figure}
		\centering
		\begin{subfigure}{0.24\linewidth}
			\includegraphics[width=\linewidth]{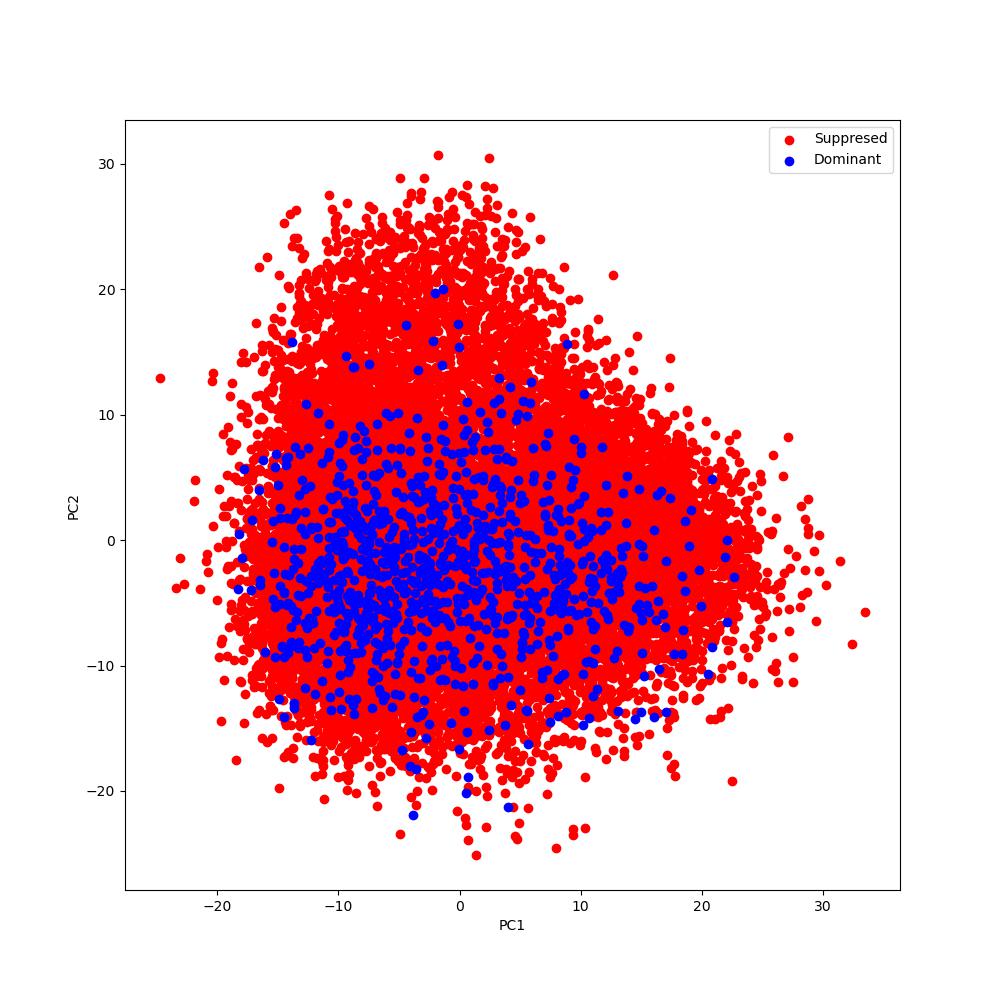}
			\caption{Feature 1}
		\end{subfigure}
		\begin{subfigure}{0.24\linewidth}
			\includegraphics[width=\linewidth]{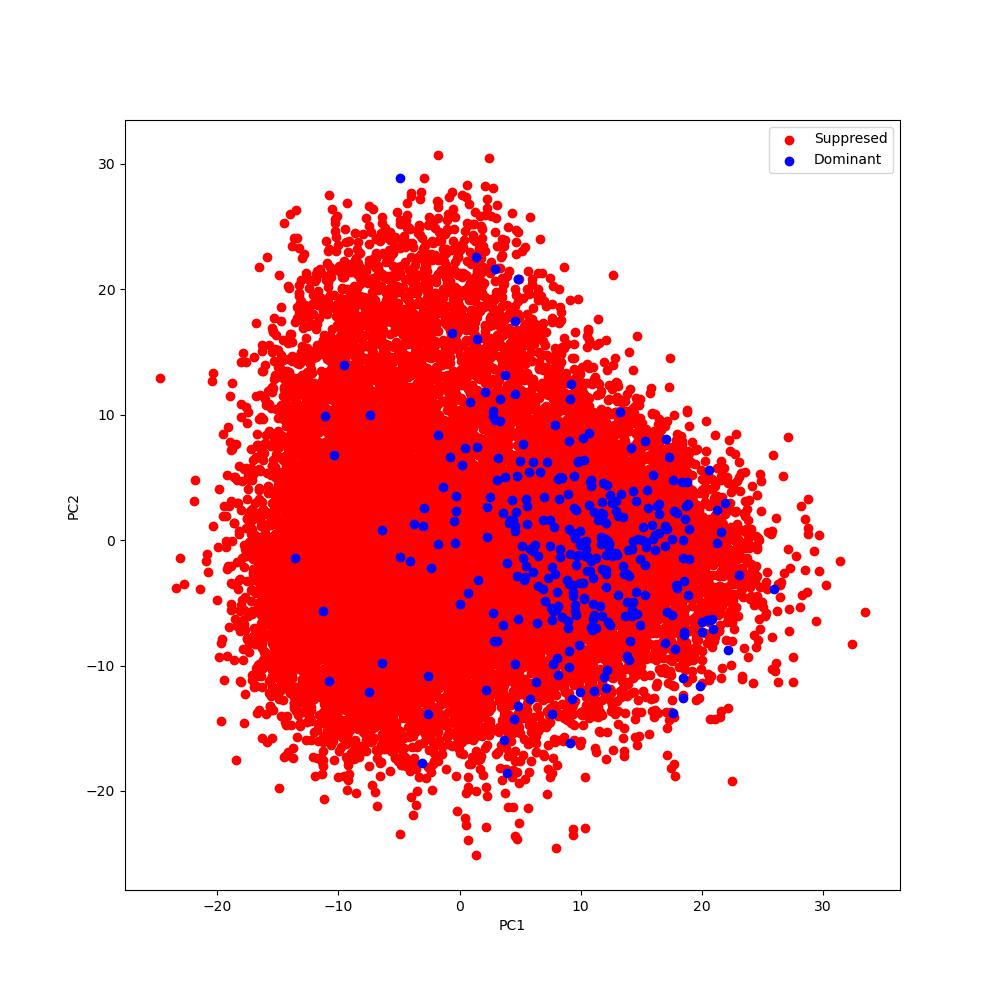}
			\caption{Feature 2}
		\end{subfigure}
		\begin{subfigure}{0.24\linewidth}
			\includegraphics[width=\linewidth]{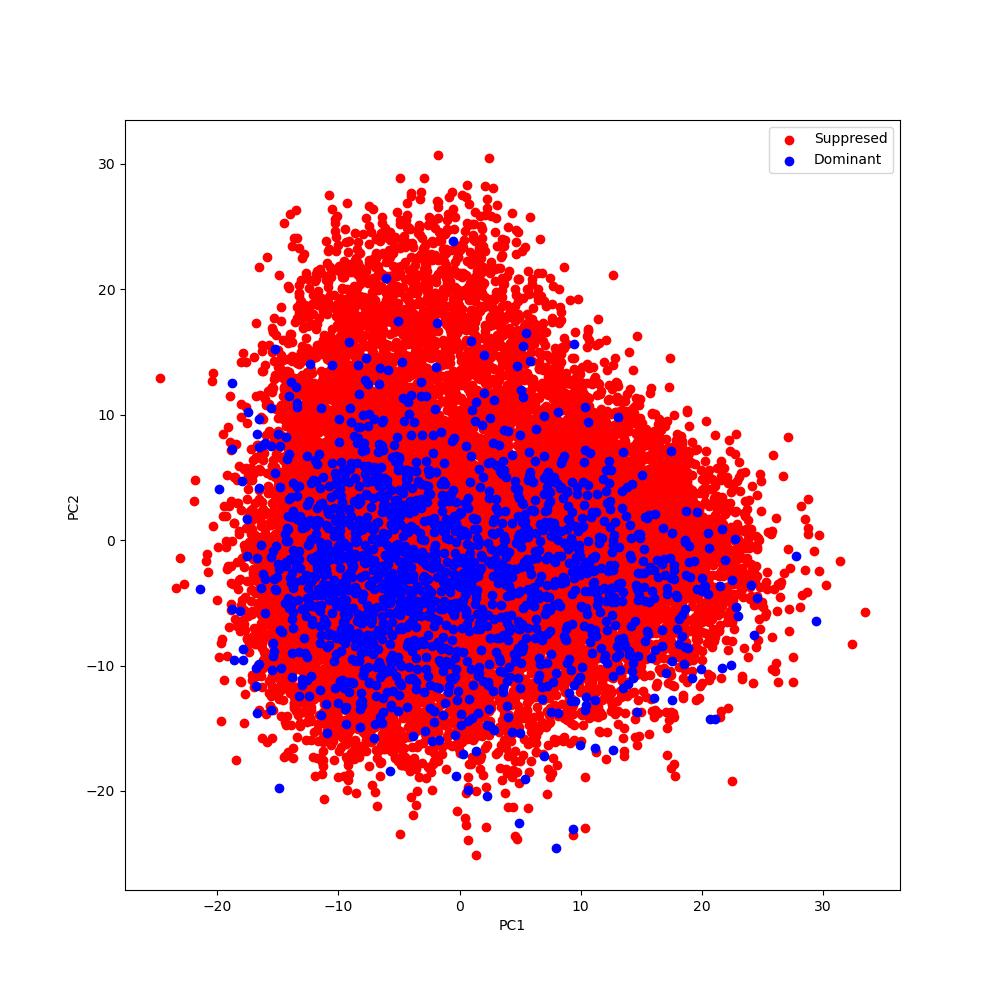}
			\caption{Feature 3}
		\end{subfigure}
		\begin{subfigure}{0.24\linewidth}
			\includegraphics[width=\linewidth]{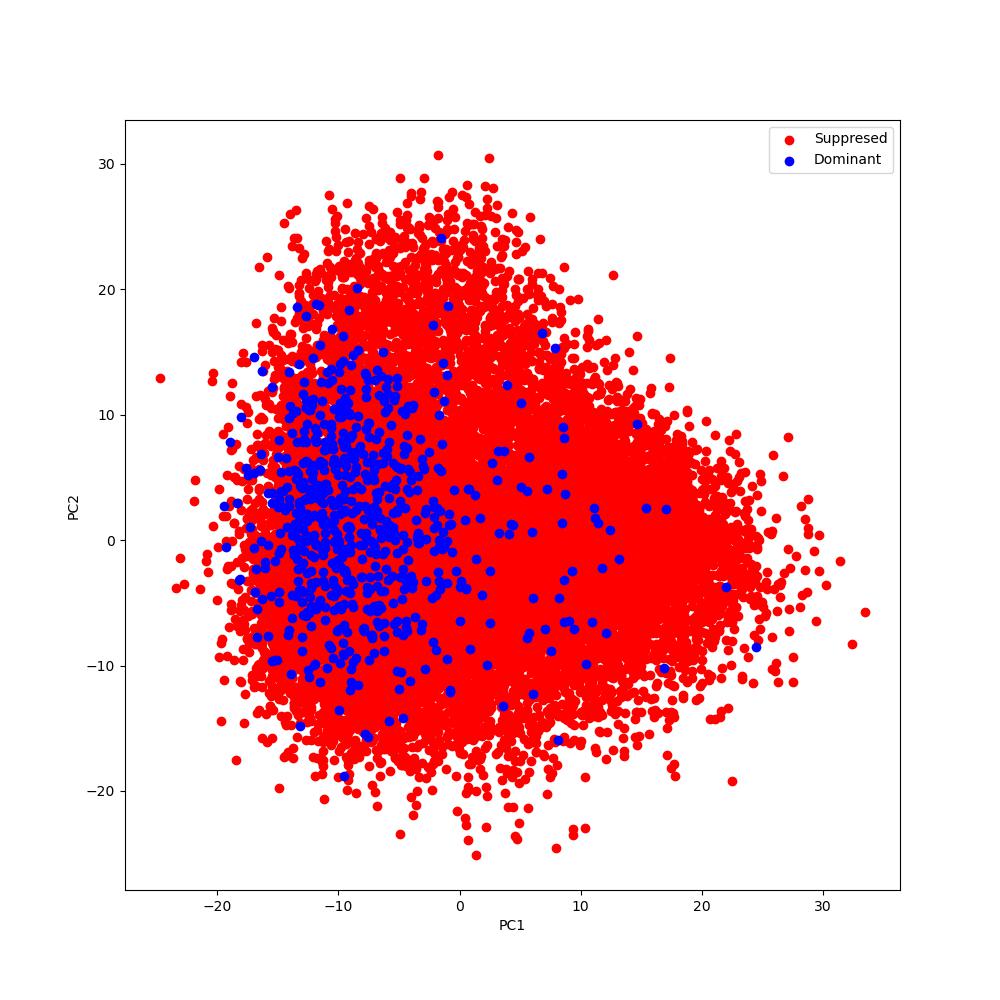}
			\caption{Feature 4}
		\end{subfigure} \\
		\begin{subfigure}{0.24\linewidth}
			\includegraphics[width=\linewidth]{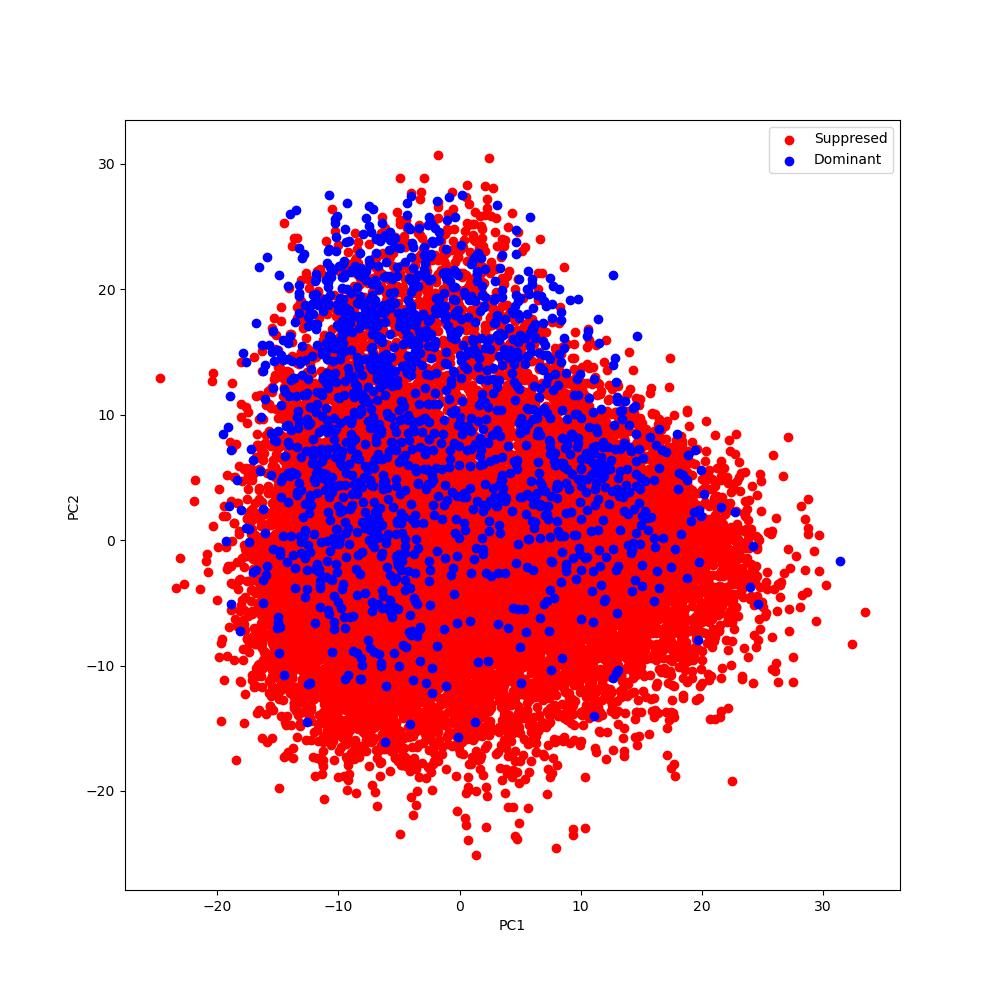}
			\caption{Feature 5}
		\end{subfigure}
		\begin{subfigure}{0.24\linewidth}
			\includegraphics[width=\linewidth]{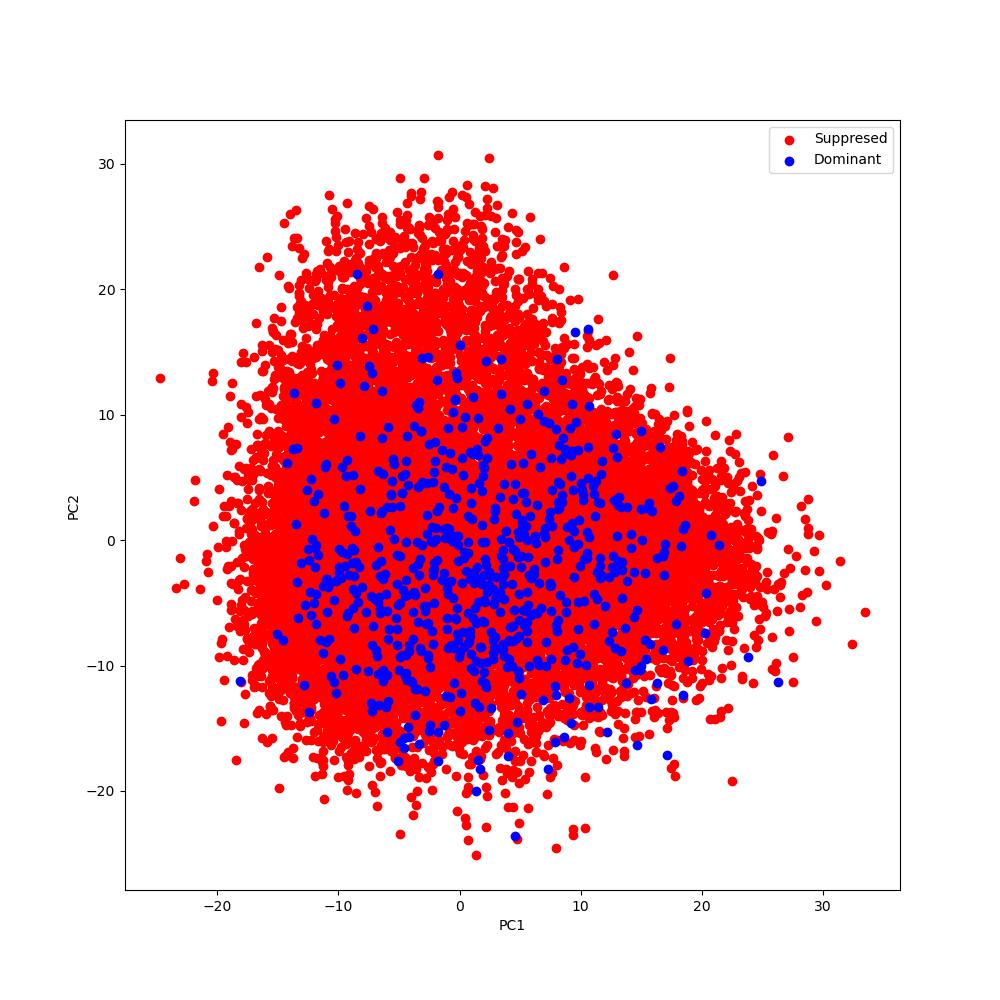}
			\caption{Feature 6}
		\end{subfigure}
		\begin{subfigure}{0.24\linewidth}
			\includegraphics[width=\linewidth]{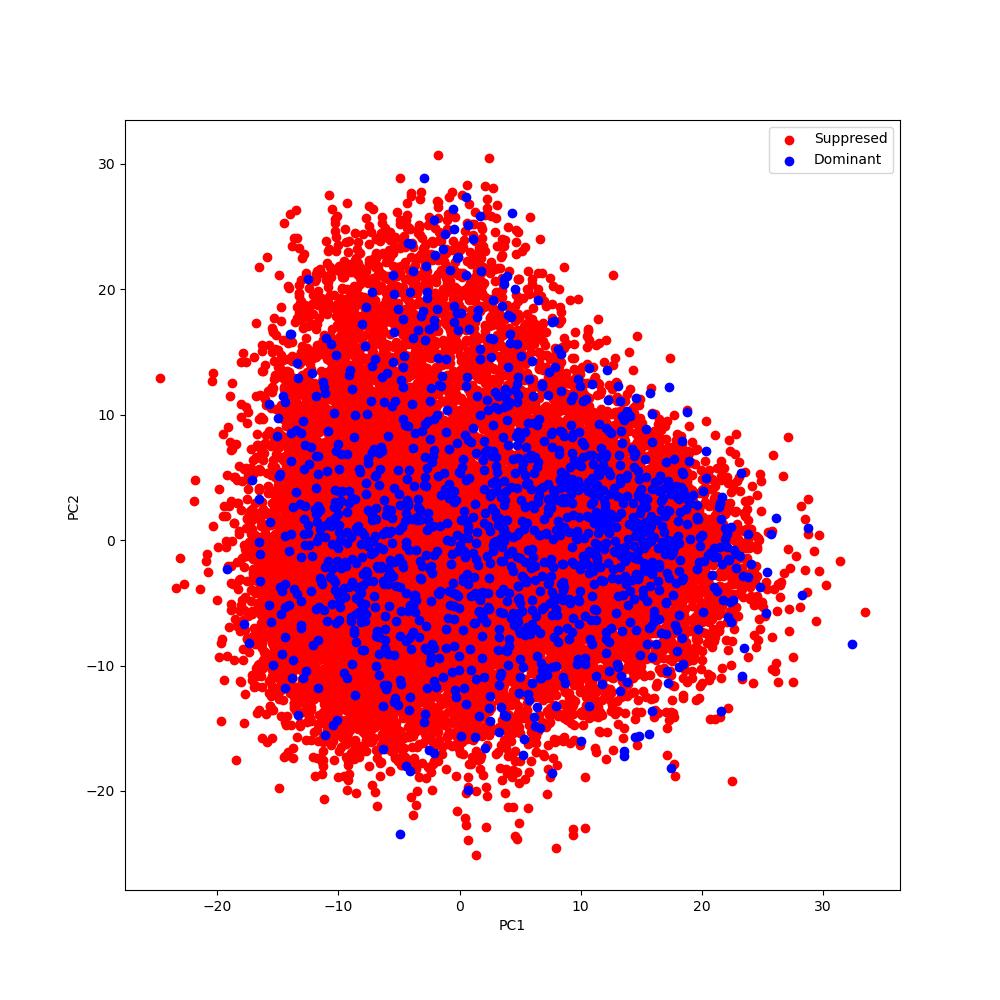}
			\caption{Feature 7}
		\end{subfigure}
		\begin{subfigure}{0.24\linewidth}
			\includegraphics[width=\linewidth]{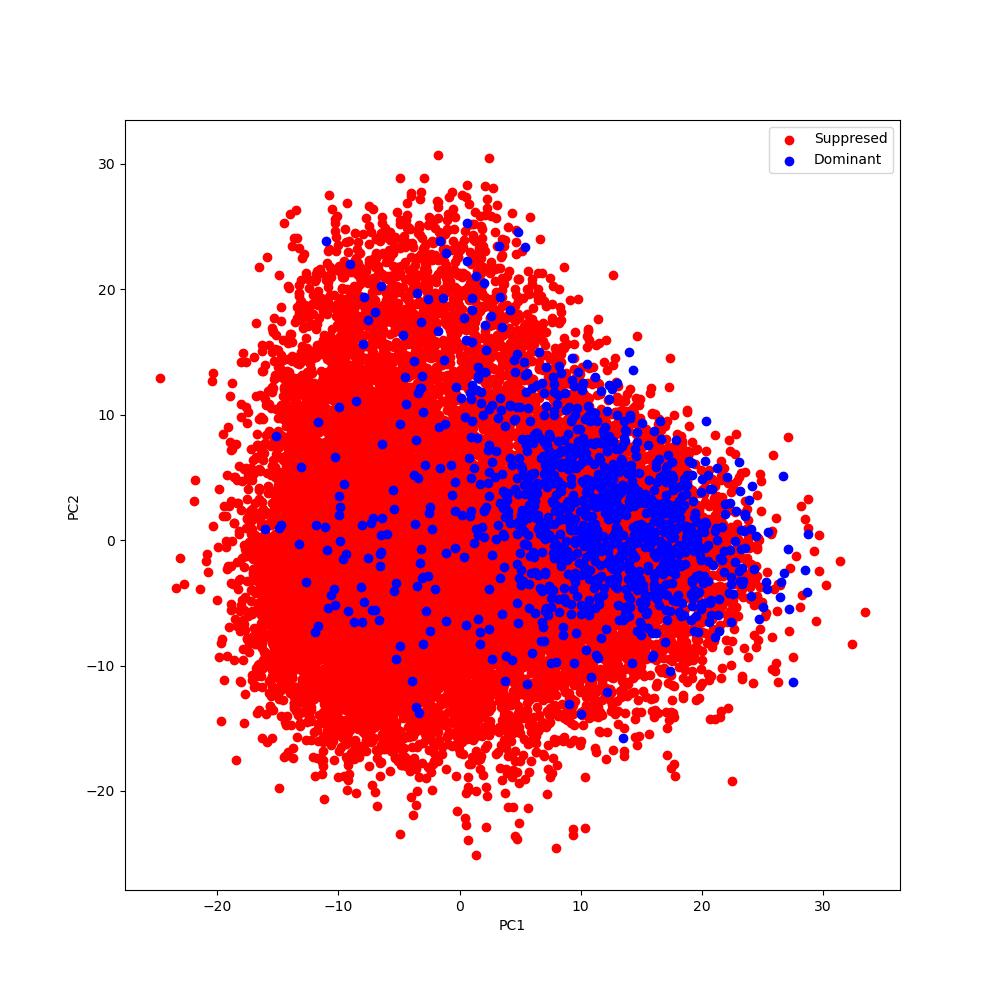}
			\caption{Feature 8}
		\end{subfigure} \\
		\begin{subfigure}{0.24\linewidth}
			\includegraphics[width=\linewidth]{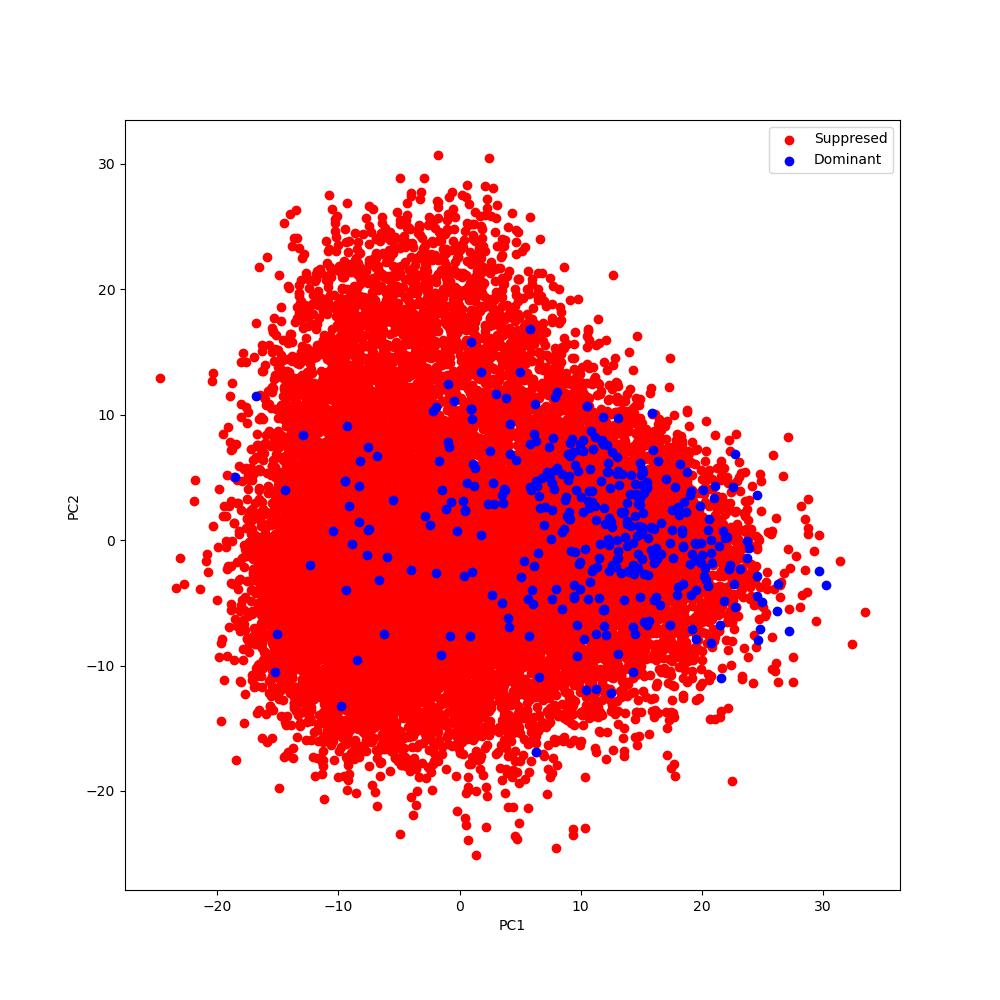}
			\caption{Feature 9}
		\end{subfigure}
		\begin{subfigure}{0.24\linewidth}
			\includegraphics[width=\linewidth]{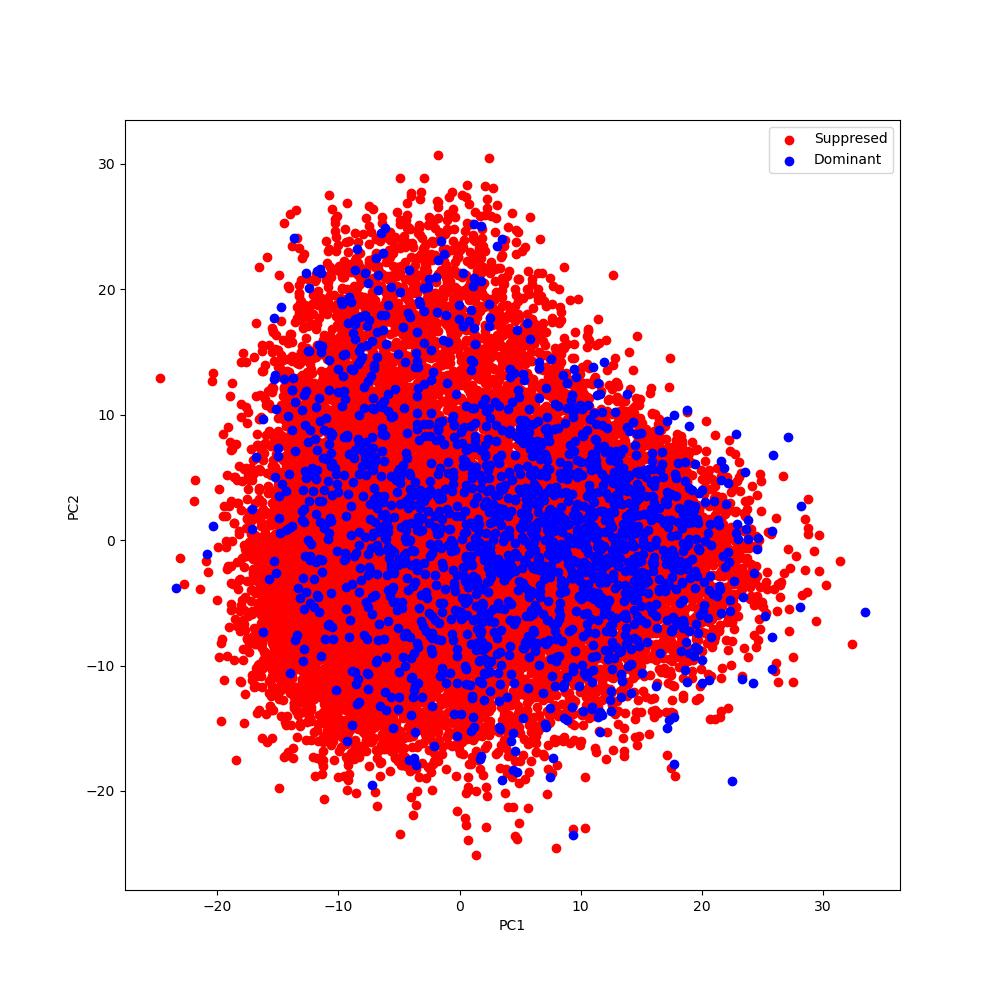}
			\caption{Feature 10}
		\end{subfigure}
		\begin{subfigure}{0.24\linewidth}
			\includegraphics[width=\linewidth]{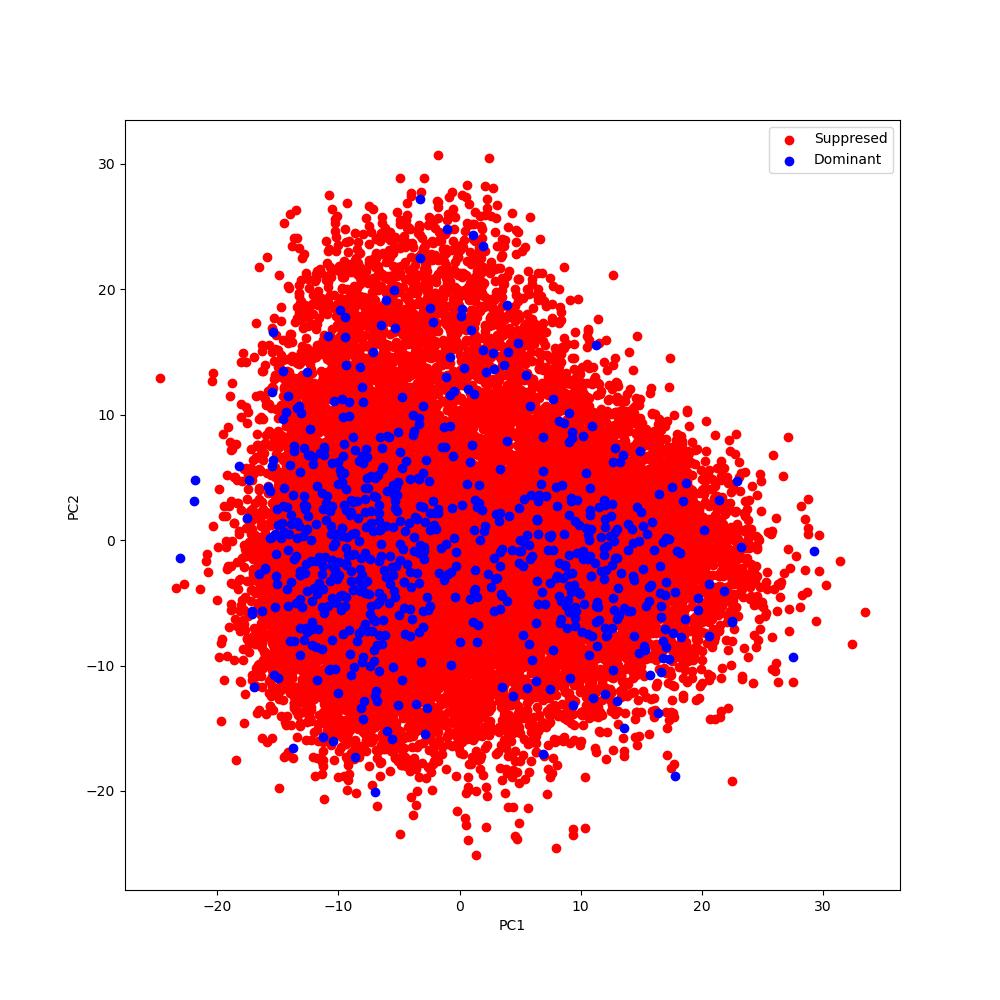}
			\caption{Feature 11}
		\end{subfigure}
		\begin{subfigure}{0.24\linewidth}
			\includegraphics[width=\linewidth]{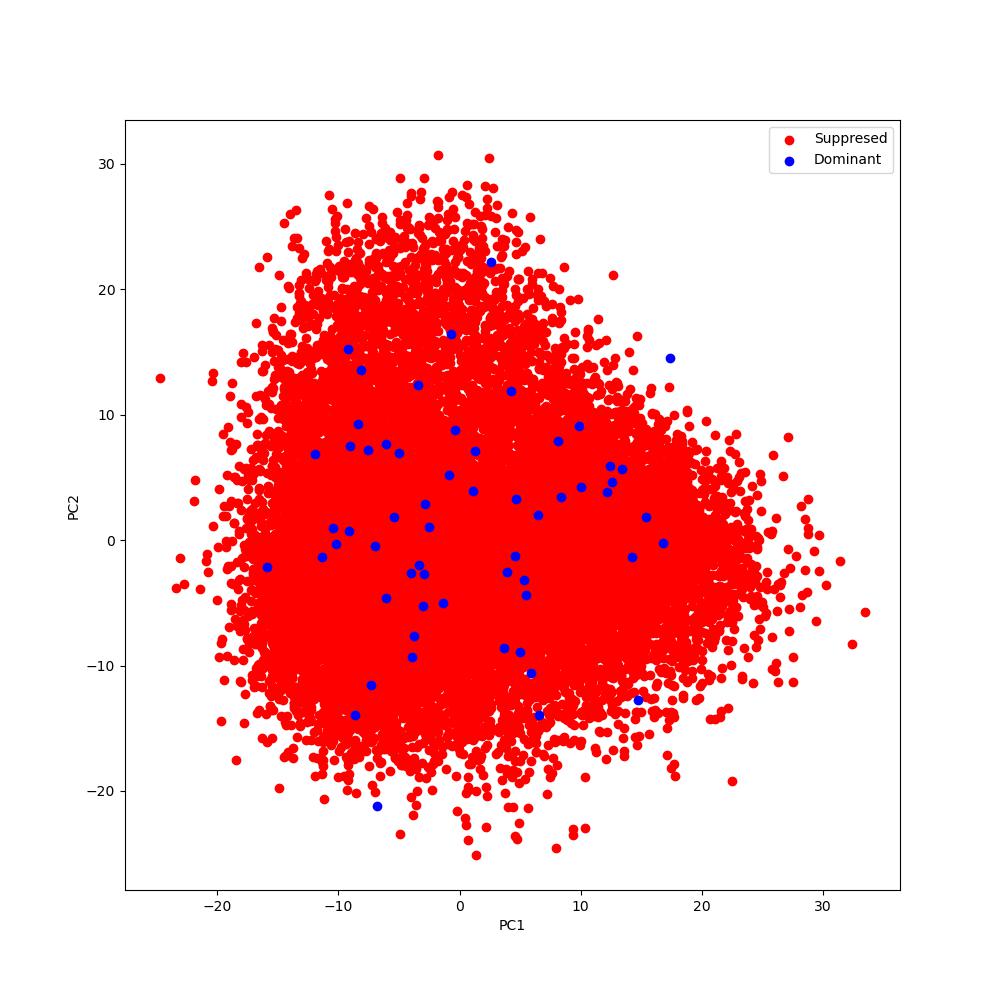}
			\caption{Feature 12}
		\end{subfigure} \\
		\begin{subfigure}{0.24\linewidth}
			\includegraphics[width=\linewidth]{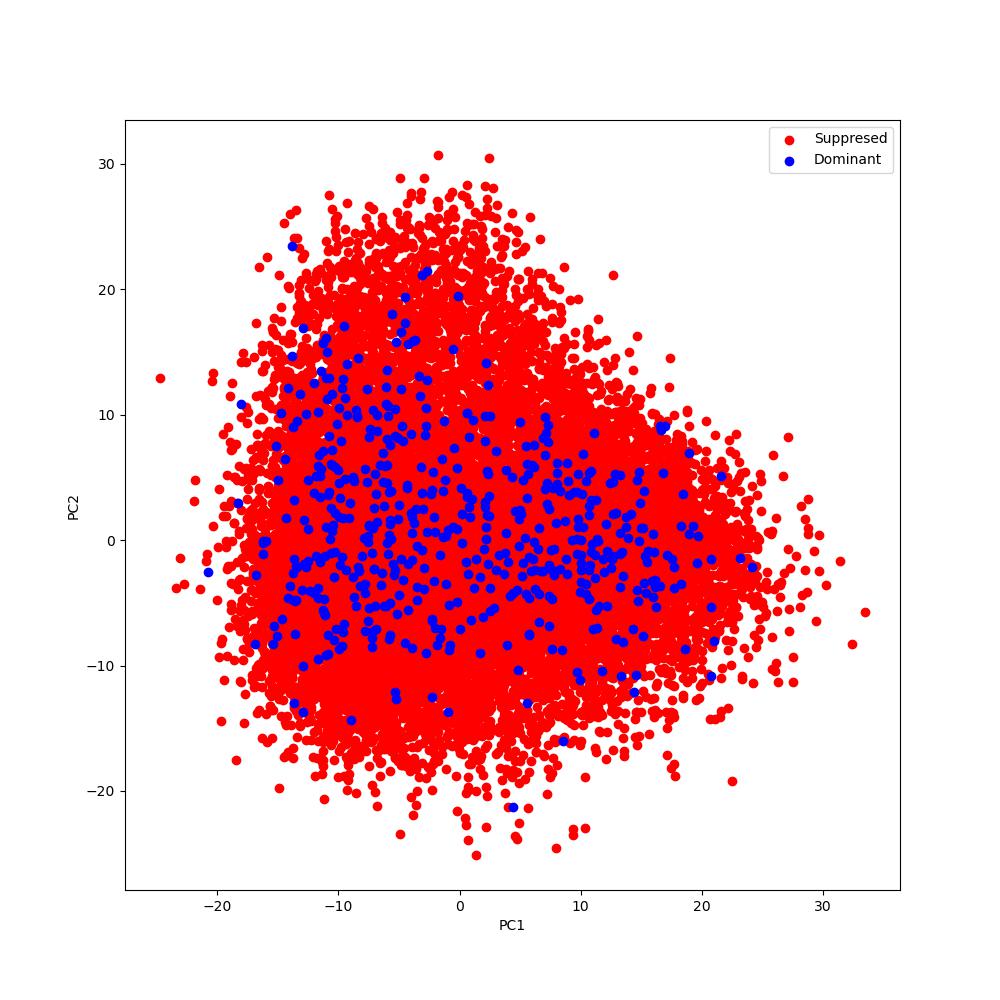}
			\caption{Feature 13}
		\end{subfigure}
		\begin{subfigure}{0.24\linewidth}
			\includegraphics[width=\linewidth]{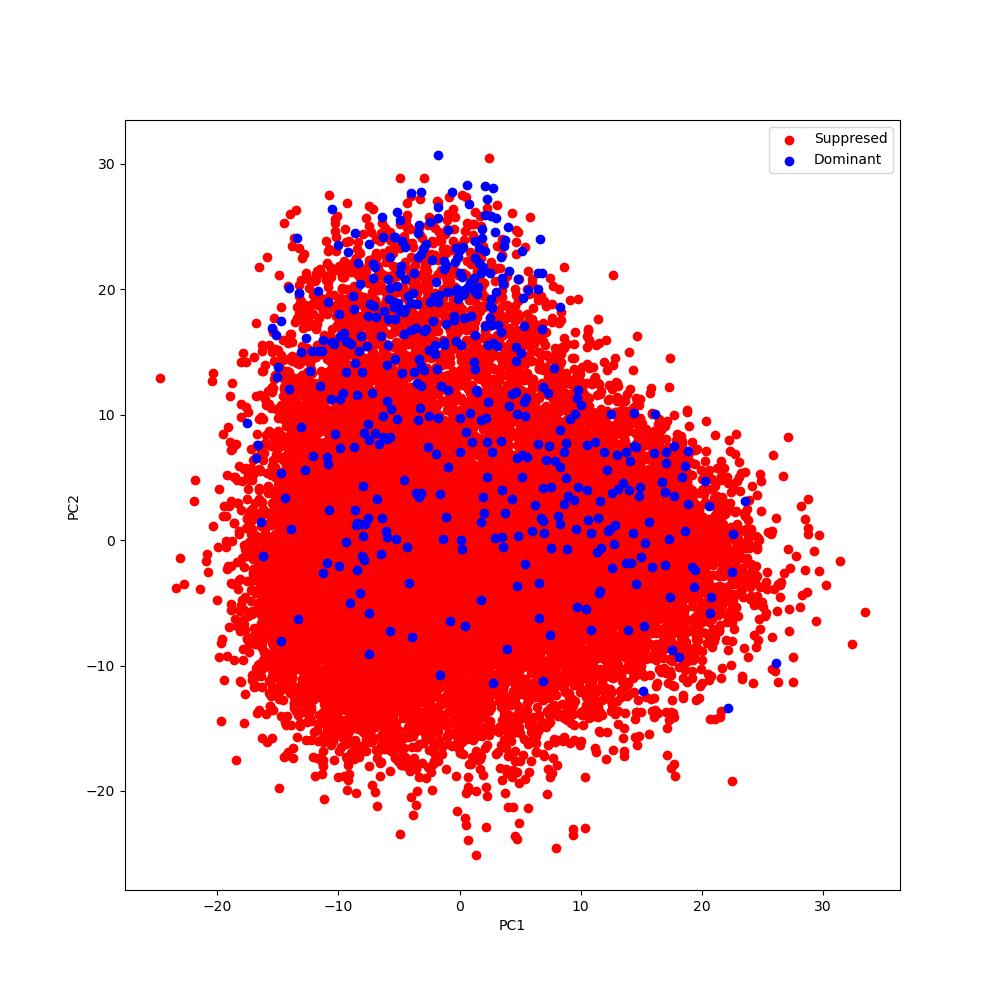}
			\caption{Feature 14}
		\end{subfigure}
		\begin{subfigure}{0.24\linewidth}
			\includegraphics[width=\linewidth]{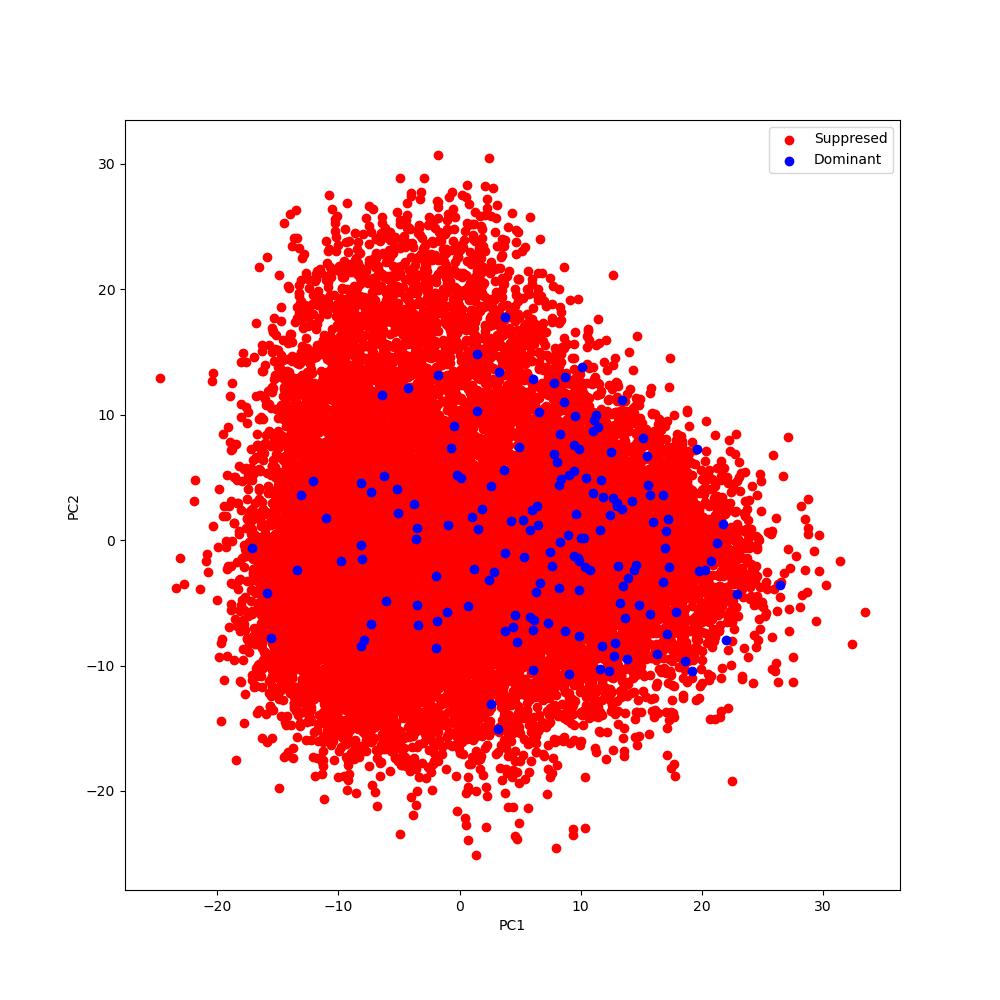}
			\caption{Feature 15}
		\end{subfigure}
		\begin{subfigure}{0.24\linewidth}
			\includegraphics[width=\linewidth]{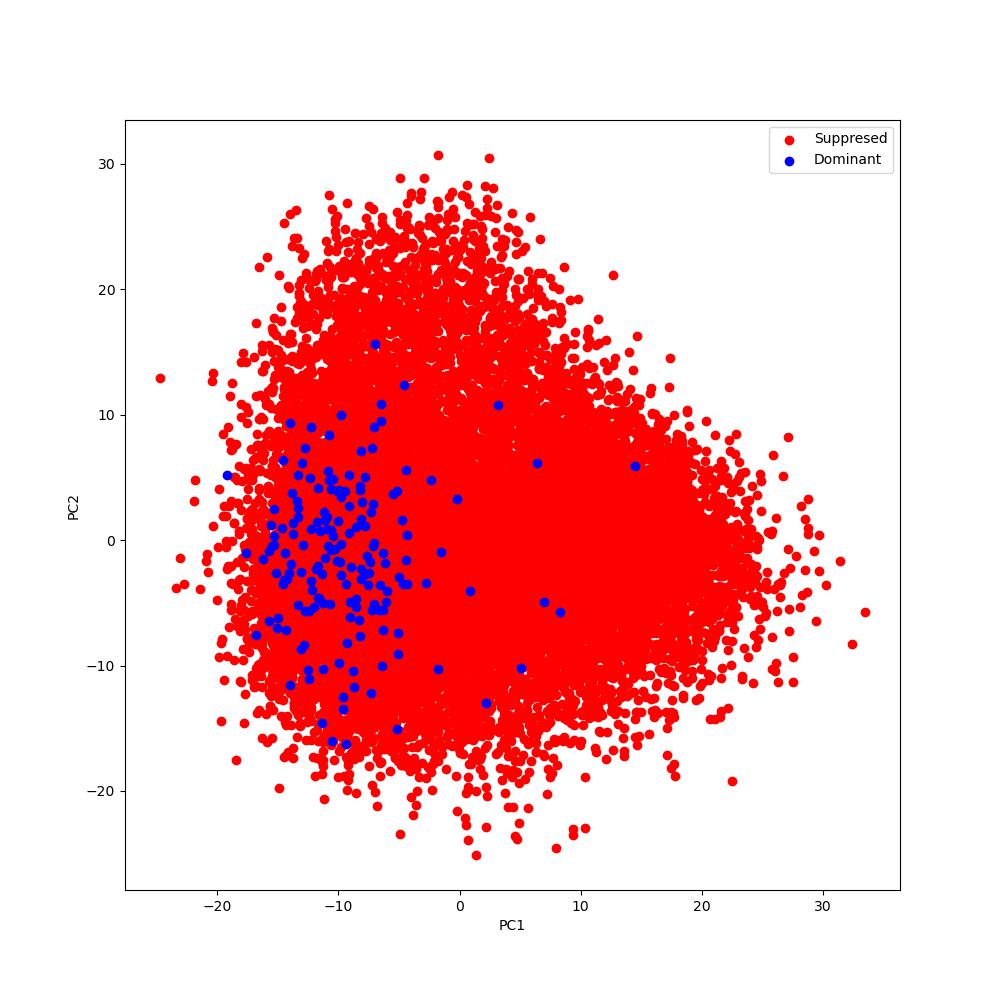}
			\caption{Feature 16}
		\end{subfigure} \\
		
		\begin{subfigure}{0.24\linewidth}
			\includegraphics[width=\linewidth]{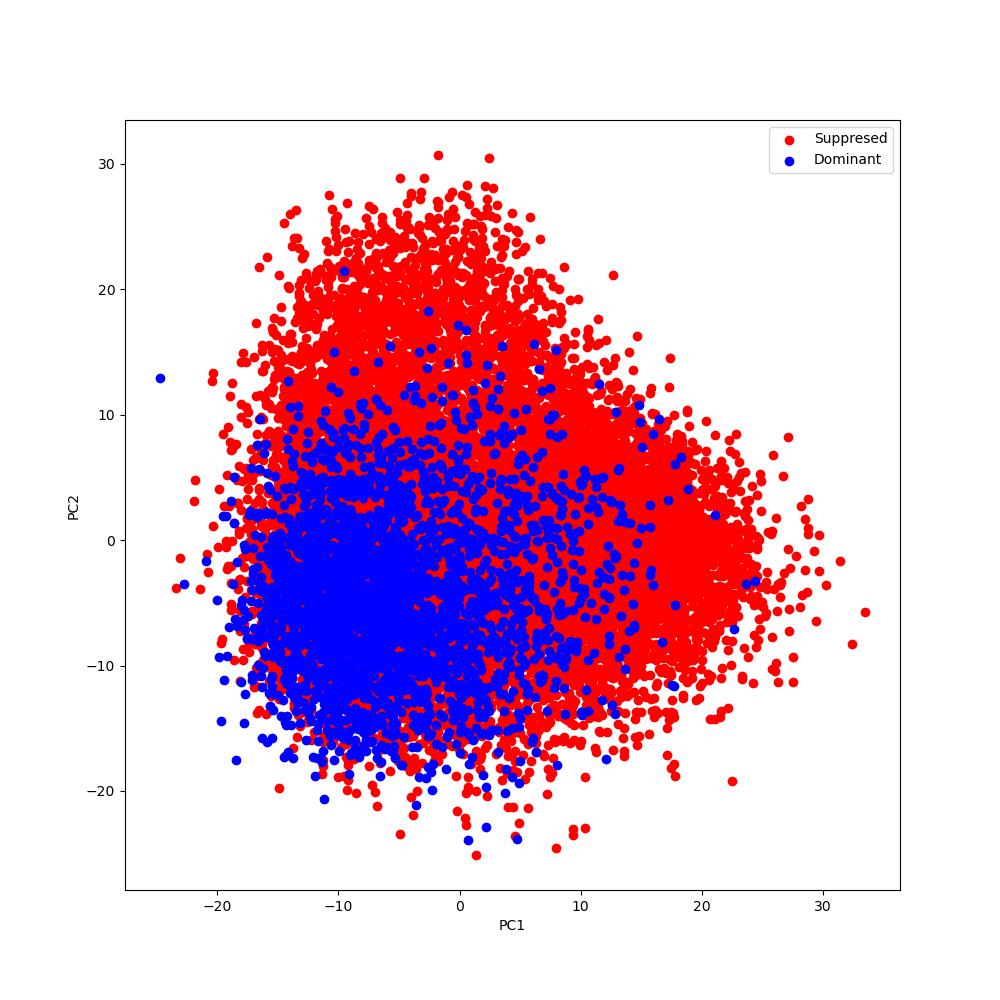}
			\caption{Feature 17}
		\end{subfigure}
		\begin{subfigure}{0.24\linewidth}
			\includegraphics[width=\linewidth]{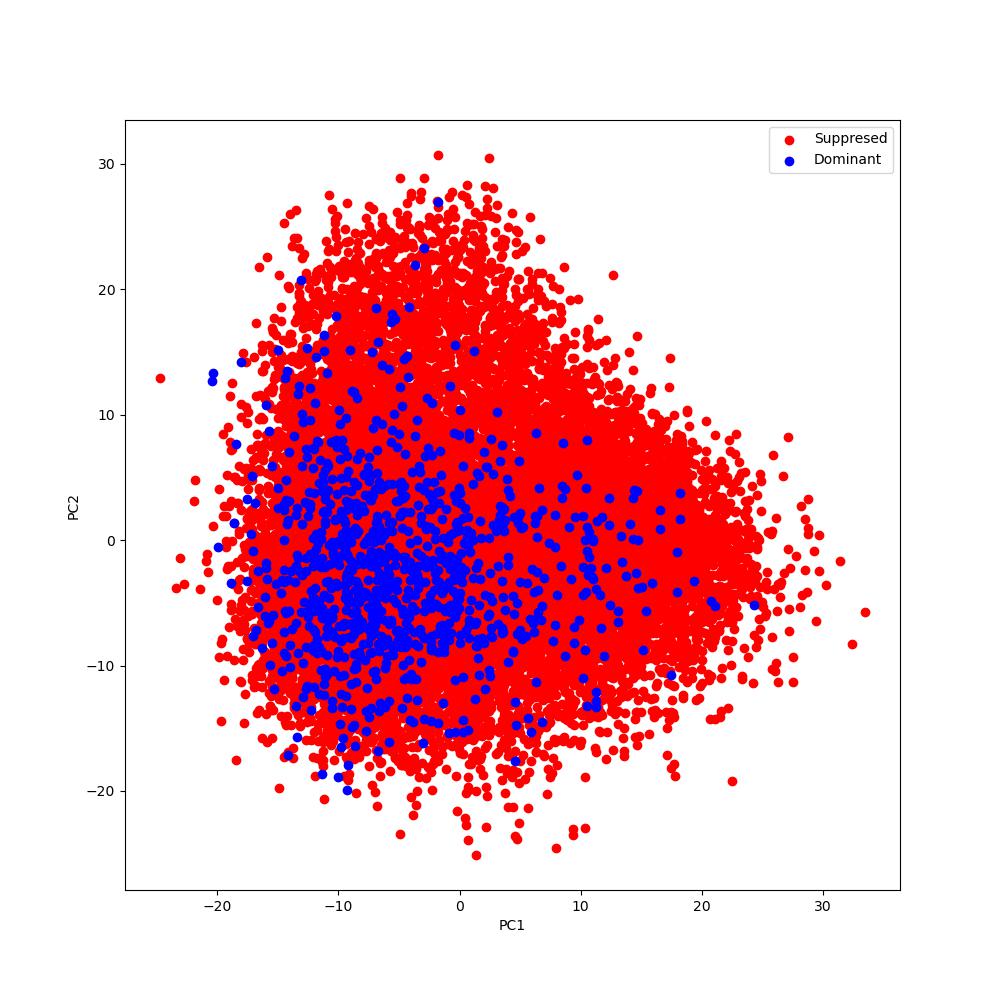}
			\caption{Feature 18}
		\end{subfigure}
		\begin{subfigure}{0.24\linewidth}
			\includegraphics[width=\linewidth]{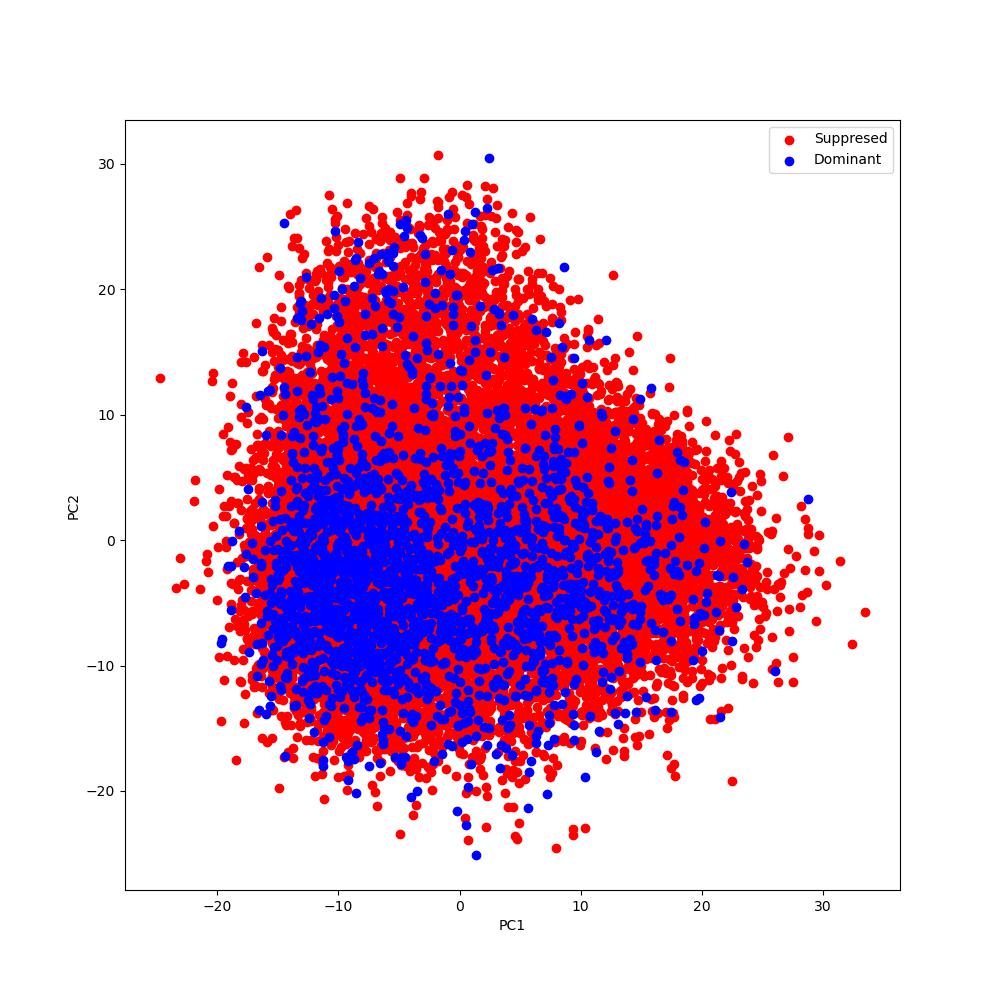}
			\caption{Feature 19}
		\end{subfigure}
		\begin{subfigure}{0.24\linewidth}
			\includegraphics[width=\linewidth]{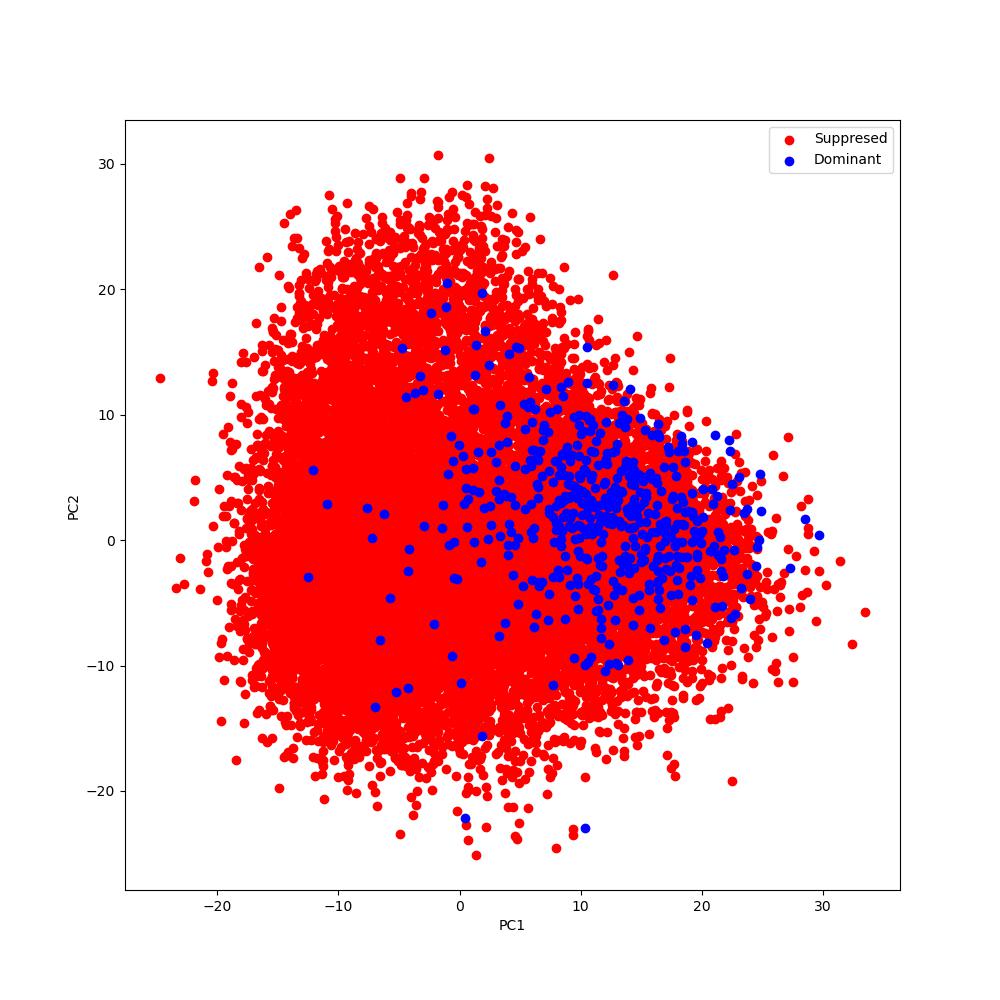}
			\caption{Feature 20}
		\end{subfigure} \\
		
		\caption{PCA projections for the deep features. Blue dots mark samples where the feature is dominant.}
		\label{fig:subfloats}
	\end{figure}

	Once we obtained the style characterization for each sample, we studied which deep  features are dominant for each sample, denoting as dominant the feature with the highest activation and how useful they are in the classification task. In order to do so, we have computed the average value and the number of times each feature presented the biggest value in a sample (Fig. \ref{fig:features_deep_plot}). Some features are clearly more dominant than others, i.e., 17 and 19, while others are remarkably low, i.e., 12. We have checked the feasibility of this task by visualizing each deep feature studied using PCA (Fig. \ref{fig:subfloats}).
	
	Then, we used a GBC to solve the classification task resulting in a $0.22$ accuracy. Since Gradient boosting is considered state-of-the-art in the standard tabular classification \cite{fumanal2023krypteia}, we can interpret this $0.22$ as an upper bound of accuracy for a FRBC. However, we can convert this problem using a One-versus-All scheme. In this way, instead of one multi-class problem, we have 20 binary classification problems. As these problems are heavily imbalanced, we use the MCC to evaluate the results for each feature. They are shown in Table \ref{tab:mcc_features_OVA}. 
	
	The lack of positive samples for each feature in comparison with the negative ones deeply affects the performance of the GBC. In order to give a more reliable estimation of the performance of the system, we randomly subsampled a balanced partition for each feature (i.e., we perform random oversampling,  Table \ref{tab:mcc_features_OVA}, column 2). Using these models, we checked the importance that they gave to each style for their predictions. Based on the relevant ones (when the importance value is bigger than the average), we use a FRBC that learns the corresponding rules to map from the input data to the desired class for each deep feature.
	
	Using this model and fuzzy linguistic variables, we can characterize each painting according to the expressiveness of their visual traits. Table \ref{tab:res_features_mcc} shows the MCC for the classification of each of the features using a FRBC. As expected from the PCA visualizations in Fig. \ref{fig:subfloats}, the performance is very different among the deep features. Finally, in Fig. \ref{tab:rules_deep_feats}, we show the results obtained for some of the features where the classification was most successful.
	
	\begin{table}[]
		\centering
		\caption{Performance measured using MCC for a GBC in the original and a balanced partition obtain by subsampling the original dataset ($-1$ is the worst possible value and $1$ the best).}
		\begin{tabular}{ccc}
			\toprule
			& \multicolumn{2}{c}{MCC performance} \\
			Feature & Original partition & Balanced partition \\
			\midrule
			1 &  0.08 & 0.26\\
			2 &  0.00& 0.52\\ 
			3 &  0.10& 0.30\\
			4 &  0.03& 0.43\\ 
			5 &  0.23& 0.48\\
			6 &  0.02& 0.25\\ 
			7 &  0.05& 0.25\\
			8 &  0.06& 0.54\\ 
			9 &  0.10& 0.40\\
			10 &  0.07& 0.21\\ 
			11 &  0.07& 0.37\\
			12 &  0.00& 0.34\\ 
			13 &  0.00& 0.18\\
			14 &  0.20& 0.56\\ 
			15 &  0.00& 0.19\\
			16 &  0.00& \textbf{0.60}\\ 
			17 &  0.22& 0.42\\
			18 &  0.05& 0.34\\ 
			19 &  0.06& 0.20\\
			20 &  0.00& 0.56\\ 
			\bottomrule
		\end{tabular}
		\label{tab:mcc_features_OVA}
	\end{table}
	
	\begin{table}[]
		\centering
		\caption{MCC obtained for each feature using a FRBC.}
		\begin{tabular}{cc| cc}
			\toprule
			Feature & MCC & Feature & MCC  \\
			\midrule
			1 & 0.2369 & 11  & 0.0000 \\ 
			2 & 0.5001 & 12  & 0.0000\\
			3 & 0.2076 & 13  & 0.0954\\
			4 & 0.2076 & 14  & 0.4666\\
			5 & 0.4440 & 15  & \textbf{0.6888}\\
			6 & 0.1611 & 16  & 0.4714\\
			7 & 0.1941 & 17  & 0.3708\\ 
			8 & 0.4512 & 18  & 0.1478\\
			9 & 0.3612 & 19  & -0.2377\\
			10 & 0.2432 & 20 & 0.4552\\
			\bottomrule
		\end{tabular}
		\label{tab:res_features_mcc}
	\end{table}
	
	\begin{figure*}[]
		\centering
		\caption{Most important rules (DS $>$0.1) that identify some of the deep features studied.}
			\begin{tabular}{l|c}
				\toprule
				\multicolumn{1}{c}{Feature 2} & DS \\
				\midrule
				
				IF Early Renaissance IS High AND Northern Renaissance IS High & 0.4968 \\
				\midrule
				\multicolumn{2}{c}{Feature 8} \\
				\midrule
				IF Early Renaissance IS High & 0.4332 \\
				\midrule
				\multicolumn{2}{c}{Feature 15} \\
				\midrule
				
				IF Cubism IS Low AND Early Renaissance IS High AND Pointillism IS Low & 0.1065 \\
				IF Early Renaissance IS High AND Rococo IS Low & 0.3906 \\
				\midrule
				\multicolumn{2}{c}{Feature 16} \\
				\midrule
				
				IF Analytical Cubism IS Low AND Naive Art Primitivism IS Medium AND Pointillism IS Medium & 0.2199 \\
				IF Contemporary Realism IS High AND Cubism IS Low AND High Renaissance IS Low & 0.2932 \\
				IF Analytical Cubism IS Low AND Color Field Painting IS Medium AND Pop Art IS Low & 0.1334\\
				
				\bottomrule
			\end{tabular}
		
		\label{tab:rules_deep_feats}
	\end{figure*}

	\section{Conclusions and future lines} \label{sec:conclusions}
	In this paper, we have proposed a new method to combine visual features and contextual annotations, using both a fuzzy membership encoding based on the FCM, a FRBC and CLIP features. We used these methods in a classification framework that considers a fine-tuned ResNet50 enriched to extract the visual features from a dataset of artistic images. This framework learns to solve a classification problem and to reconstruct the features extracted from the contextual information for each image, which helps the network generalize better, as it does not need to rely only on visual cues to classify each sample. Besides, we have introduced different XAI methods using fuzzy rules to interpret the features and results obtained with these methods.
	
	The comparison between context-aware models with similar visual-only classification frameworks shows favorable results for the former ones, as originally expected. We obtained the best results overall using an MTL paradigm with contextual information. Using fuzzy rules, we also showed how some of the deep features used by the best model can be characterized according to the relevant parts of the image and the style of painting. In addition, we reported how some painters can be successfully distinguished one from another using fuzzy rules and painting styles. As a way of example, we show in this paper a comparison between Paul Gauguin and Vincent Van Gogh
	
	Future lines of our research shall study more expressive features to represent some of the image characteristics \cite{feats} and apply methods to improve the performance of the FRBC in imbalanced datasets. We also intend to develop a metric that can compute how good is a commentary that describes an image so that the additional information present in the text can be quantified.
	
	\section{Acknowledgements}
	This work was supported in part by Oracle Cloud credits and related resources provided by Oracle for Research and by MCIN/AEI/10.13039/501100011033 and ERDF “A way of making Europe” under grant CONFIA (PID2021-122916NB-I00).

	\bibliographystyle{IEEEtran}
	\bibliography{sn-bibliography}

\end{document}